\documentclass{article}

\PassOptionsToPackage{numbers,sort&compress}{natbib}
\usepackage[eandd,final]{neurips_2026}

\usepackage[utf8]{inputenc}
\usepackage[T1]{fontenc}
\usepackage{hyperref}
\hypersetup{hidelinks,pdftitle={Prefix Executability: Evaluating Tool-Using Agents Beyond Final Success},pdfauthor={Amir H. Rezaeian, Weiyi Sun, Yassine Benajiba, Dan Roth}}
\usepackage{xurl}
\usepackage{booktabs}
\usepackage{amsmath,amssymb,amsfonts}
\usepackage{microtype}
\usepackage{xcolor}
\usepackage{enumitem}
\usepackage{graphicx}
\usepackage{mathtools}
\usepackage{tabularx}
\usepackage{etoolbox}
\usepackage{placeins}

\title{Prefix Executability: Evaluating Tool-Using Agents Beyond Final Success}

\author{%
  Amir H. Rezaeian\thanks{Correspondence: \texttt{Amir.Rezaeian@oracle.com}.}\hphantom{$^{\ast}$}, Weiyi Sun, Yassine Benajiba, Dan Roth\\
  Oracle AI%
}

\newcommand{\inputsection}[1]{%
  \IfFileExists{sections/#1.tex}{\input{sections/#1}}{\input{#1}}%
}

\newcommand{\rev}[1]{{#1}}
\newenvironment{revision}{\par\begingroup}{\par\endgroup}

\begin{document}
\maketitle

\begin{abstract}
Tool-use benchmarks often score agents by final task success, but terminal outcomes can obscure whether the intermediate tool-call trajectory was executable. We propose \emph{prefix executability} as a trajectory-level evaluation protocol for tool-using agents. The protocol measures whether every prefix of a predicted tool-call sequence remains executable under tool schemas, state preconditions, and data dependencies, and summarizes reliability using Prefix Executability Curves (PEC), First-Error Depth (FED), Trajectory Executability Rate (TER), horizon-normalized AUC-PEC, and Recovery Rate (RR). We instantiate the protocol on stratified samples from BFCL-v3 multi-turn and ToolSandbox\rev{, and on the complete Airline/Retail base splits of $\tau^3$-bench}. \rev{We observe that} outcome-only and prefix-level rankings diverge: agents whose execution is conditioned on an unverified plan often achieve higher point-estimate final success than non-plan-conditioned baselines while substantially lowering TER and AUC-PEC. These divergences are precisely the failure modes that prefix executability is designed to expose: they reveal whether apparent success rests on an executable trajectory, where invalidity first enters, and whether success follows recovery from an earlier failure. As a validation intervention, explicit pre-execution plan checking \rev{can improve} executable-prefix survival while maintaining competitive terminal success. These results show that final success alone is insufficient for evaluating tool-use agents and that prefix-level executability provides an actionable reliability signal for comparing agents, diagnosing failures, and assessing validation mechanisms.

\end{abstract}

\section{Introduction}
Tool-using language agents operate over APIs, user-defined tools, and interactive software environments where correctness depends on more than a final answer. A valid run must maintain an executable sequence of intermediate actions while respecting tool schemas, cross-step dependencies, and evolving environment state~\citep{toolformer, react, bfcl, toolsandbox}. This requirement is especially acute in multi-turn settings, where an early invalid call can corrupt later state or force the agent into recovery behavior that is invisible to outcome-only scoring~\citep{bfcl, toolsandbox, nestful, taubench}. Consider a service workflow in which an agent first retrieves a record, then passes the returned identifier to update, payment, or notification tools. A later call may be syntactically well formed but non-executable if it references a stale identifier, skips a required precondition, or applies an update before the corresponding state exists; a subsequent repair can still produce the desired final state while hiding where the trajectory first became invalid.

Recent evaluation work has moved beyond simple terminal accuracy. T-Eval decomposes tool use into step-wise subskills~\citep{teval}; ToolSandbox \rev{evaluates task-specified intermediate and final conditions and checks for prohibited states or events}~\citep{toolsandbox}; TRAJECT-Bench scores tool selection, argument correctness, and dependency ordering over full trajectories~\citep{traject_bench}; ToolPRMBench and AgentProcessBench \rev{assess local process quality}~\citep{toolprmbench, agentprocessbench}; and procedure-aware evaluation shows that nominally successful runs can still violate task procedures~\citep{procedure_aware}. \rev{TRACE adds reference-free assessment of efficiency, hallucination, and adaptivity over tool-use trajectories~\citep{trace}.} Robustness studies further probe query perturbations, tool-set changes, and injected execution faults~\citep{robustness_fc, beyond_success_rate}. These efforts make a remaining gap clear: standard metrics still rarely quantify how long a naturally generated rollout remains executable before its first invalid prefix, or how often final success is achieved only after entering a failure regime.

\rev{Prefix executability complements these evaluations by checking the execution feasibility of the actual emitted rollout under explicit tool and state semantics. It credits alternative executable action sequences without requiring one gold path and records first failure, survival, and recovery. Appendix~\ref{app:protocol-comparison} compares what prior evaluation protocols measure.}

We study \emph{prefix-level executability failures}: trajectory prefixes that may look plausible as text but are non-executable under the available tool schema, explicit preconditions, cross-step references, or current environment state. We argue that prefix executability is a missing reliability axis because final success and reference-trajectory match conflate three questions: how long the rollout remains valid, where validity first breaks, and whether later actions recover from the break. We therefore formalize executable prefixes and report a metric family centered on Prefix Executability Curves (PEC), First-Error Depth (FED), horizon-normalized AUC-PEC, Trajectory Executability Rate (TER), and Recovery Rate (RR).  These metrics primarily diagnose reliability; when reported alongside trajectory
length, latency, or monetary cost, they also support efficiency-sensitive
analysis by distinguishing short, valid rollouts from long or failure-prone ones.

These distinctions are operationally important in cost-sensitive tool use. In workflows that trigger external transactions, consume scarce resources, modify persistent state, or require human escalation, the depth of the first invalid action can determine both the cost of failure and the feasibility of recovery. Prefix executability therefore provides a practical evaluation signal for deciding not only whether an agent eventually succeeds, but also whether \rev{execution remains uninterrupted or requires recovery} along the way.

The paper is evaluation-first. We deliberately use a minimal Dual-Loop planner-verifier intervention only as a controlled probe: \rev{we test whether pre-execution validation improves those metrics even when final success rankings tell a different story}. This framing separates the evaluation contribution from the particular intervention and lets us ask when plan conditioning helps tool use, and when it amplifies invalid intermediate structure.

Our contributions are:
\begin{enumerate}[leftmargin=*, nosep]
    \item We formalize executable prefixes for tool-using agents and define PEC, FED, TER, AUC-PEC, and RR as a prefix-level evaluation protocol for variable-horizon trajectories. \rev{Appendix~\ref{app:theory} establishes exact properties and consistency under explicit sampling assumptions.}
    \item We instantiate the protocol on BFCL v3 multi-turn and ToolSandbox with stratified 200-example samples\rev{, and on all 164 Airline/Retail base-split tasks in $\tau^3$-bench}, \rev{using} shared task IDs across methods and models, and task-level confidence intervals.
    \item We show that outcome-only and prefix-level rankings can diverge: unverified plan conditioning can improve final success while sharply reducing executability, whereas pre-execution validation consistently improves TER, AUC-PEC, and early PEC across three LLM conditions \rev{on BFCL v3 and ToolSandbox}. A representative BFCL setting illustrates the pattern: an unverified plan-conditioned executor can reach the correct terminal outcome while emitting invalid intermediate tool-call prefixes, whereas Dual-Loop preserves terminal performance while restoring prefix executability. \rev{The $\tau^3$-bench audit further demonstrates that prefix reliability and terminal success can rank methods differently.}
\end{enumerate}

\rev{We evaluate BFCL v3 and ToolSandbox because} together they cover multi-step function composition and stateful, on-policy tool interaction~\citep{bfcl, toolsandbox}. \rev{The Airline/Retail audit evaluates the protocol in policy-constrained user--agent--environment interaction in $\tau^3$-bench. Together, the three benchmarks cover 564 task instances per model--method condition and 6,768 rollouts.}

\section{Prefix-Level Executability Evaluation}

\subsection{Executable Prefixes}

We model an agent rollout as a predicted trajectory $\hat{\tau}=(a_1,\ldots,a_T)$, where each ordinary action $a_t$ is a tool invocation with arguments. Let $s_t$ denote the environment state immediately before executing $a_t$. An action is executable if the selected tool is available, the arguments conform to its schema, all explicit preconditions hold in $s_t$, and any referenced intermediate values are present and type-consistent.

A prefix $\hat{\tau}_{\leq k}=(a_1,\ldots,a_k)$ is executable if every action in the prefix executes successfully under the states induced by prior actions. This definition evaluates feasibility rather than agreement with a single reference trajectory, which matters because many tool tasks admit multiple valid execution orders or alternative executable call sequences~\citep{bfcl, toolsandbox, taubench}.

\rev{The benchmark-side scorer is deterministic and uses no LLM judge. It checks exposed tools, argument binding and validators, current and type-compatible references, information availability, explicit preconditions, and benchmark acceptance of the transition. Accepted actions advance state; the first rejection fixes FED. Prefix replay then stops, while the full rollout is retained for final success and recovery. This scorer is distinct from the LLM-based pre-execution verifier. Appendix~\ref{app:checker-procedure} specifies the procedure and benchmark-specific semantics.}

\subsection{Prefix-Level Reliability Metrics}
\label{sec:prefix-metrics}

For rollout $i$, let $T_i$ denote the number of evaluated tool-action slots and let $Y_i\in\{0,1\}$ indicate benchmark-defined final task success. For ordinary trajectories, $T_i$ is the number of predicted tool invocations. If an agent terminates before emitting any tool call in a selected tool-use task, the instrumentation represents this as a length-one terminal no-tool action; that action is executable only if the benchmark permits task completion without tool use. Thus $T_i\geq 1$ for all evaluated rollouts. We use $T$ and $Y$ without subscripts for the corresponding random variables, and hats denote finite-sample estimates over the evaluated rollouts.

The first primitive is \emph{First-Error Depth} (FED):
\begin{equation}
\mathrm{FED}_i = \min\{k\in\{1,\ldots,T_i\}: \hat{\tau}_{i,\leq k}\text{ is not executable}\},
\label{eq:fed}
\end{equation}
with $\mathrm{FED}_i=T_i+1$ if every prefix through the full trajectory is executable. Hence, for any $k\leq T_i$,
\begin{equation}
\hat{\tau}_{i,\leq k}\text{ is executable}\quad\Longleftrightarrow\quad \mathrm{FED}_i>k.
\label{eq:fed-prefix}
\end{equation}

We define the main \emph{Prefix Executability Curve} (PEC) as a fixed-cohort survival curve over first-error depth,
\begin{equation}
\mathrm{PEC}(k)=\Pr[\mathrm{FED}>k]=\Pr[T\geq k\ \wedge\ \hat{\tau}_{\leq k}\text{ is executable}].
\label{eq:pec}
\end{equation}
This is a discrete survival function in the usual time-to-event sense~\citep{survival_textbook, kaplan_meier}. We do not apply a censoring correction: a rollout that terminates before depth $k$ is not unobserved at depth $k$, but instead lies outside the event $T\geq k$ in Eq.~\eqref{eq:pec}. The empirical estimate over $N$ evaluated rollouts is
\begin{equation}
\widehat{\mathrm{PEC}}(k)
=\frac{1}{N}\sum_{i=1}^{N}\mathbf{1}\{\mathrm{FED}_i>k\}
=\frac{1}{N}\sum_{i=1}^{N}\mathbf{1}\{T_i\geq k\ \wedge\ \hat{\tau}_{i,\leq k}\text{ is executable}\}.
\label{eq:pec-estimator}
\end{equation}
The denominator is fixed at $N$, the number of rollouts for a given dataset, model, and agent variant. Therefore $\widehat{\mathrm{PEC}}(k)$ is monotone non-increasing in $k$; the rollout horizon $T_i$ should not be confused with the sample size $N$.

For diagnostics, we also track the eligible-set size
\begin{equation}
n_k=\sum_{i=1}^{N}\mathbf{1}\{T_i\geq k\},
\label{eq:eligible-size}
\end{equation}
and, when $n_k>0$, the conditional eligible-depth quantity
\begin{equation}
\widehat{\mathrm{PEC}}_{\mathrm{elig}}(k)=\frac{1}{n_k}\sum_{i=1}^{N}\mathbf{1}\{T_i\geq k\ \wedge\ \mathrm{FED}_i>k\}.
\label{eq:eligible-pec}
\end{equation}
This conditional curve answers a different question: among rollouts with at least $k$ tool-action slots, what fraction have an executable prefix of length $k$? Because the eligible set changes with $k$, $\widehat{\mathrm{PEC}}_{\mathrm{elig}}(k)$ need not be monotone and is not the survival curve in Eq.~\eqref{eq:pec}. We use Eq.~\eqref{eq:pec-estimator} for the main PEC plots and tables.
\begin{revision}
The two curves satisfy the exact support decomposition
\begin{equation}
\widehat{\mathrm{PEC}}(k)=\frac{n_k}{N}\widehat{\mathrm{PEC}}_{\mathrm{elig}}(k),\qquad n_k>0.
\label{eq:pec-support}
\end{equation}
Thus fixed-cohort PEC measures the fraction of all started rollouts that reach depth $k$ with an executable prefix; its decline can reflect either a prefix failure or termination before that depth. Eligible-depth PEC isolates validity conditional on reaching $k$.
\end{revision}

The \emph{Trajectory Executability Rate} (TER) is the probability that the full predicted trajectory is executable:
\begin{equation}
\mathrm{TER}=\Pr[\mathrm{FED}=T+1],\qquad
\widehat{\mathrm{TER}}=\frac{1}{N}\sum_{i=1}^{N}\mathbf{1}\{\mathrm{FED}_i=T_i+1\}.
\label{eq:ter}
\end{equation}
When all episodes share a common fixed horizon $T_0$, TER equals $\mathrm{PEC}(T_0)$. With variable-horizon rollouts, Eq.~\eqref{eq:ter} is the unambiguous endpoint definition.

The \emph{Recovery Rate} (RR) is final success conditional on having experienced at least one invalid prefix,
\begin{equation}
\mathrm{RR}=\Pr[Y=1\mid \mathrm{FED}\leq T],\qquad
\widehat{\mathrm{RR}}=
\frac{\sum_{i=1}^{N}\mathbf{1}\{Y_i=1\}\mathbf{1}\{\mathrm{FED}_i\leq T_i\}}
{\sum_{i=1}^{N}\mathbf{1}\{\mathrm{FED}_i\leq T_i\}},
\label{eq:rr}
\end{equation}
when the denominator is nonzero; otherwise RR is undefined for that condition. We report this denominator as RR-Eligible. RR is closest in spirit to the Recovery Rate reported by Beyond Success Rate, but conditions on first-prefix failures observed organically in the rollout rather than externally injected API faults~\citep{beyond_success_rate}.

Finally, to compare variable-length episodes with a scalar, we report a horizon-normalized area under the executable-prefix profile:
\begin{equation}
\mathrm{AUC\mbox{-}PEC}
=\mathbb{E}\left[\frac{1}{T}\sum_{k=1}^{T}\mathbf{1}\{\mathrm{FED}>k\}\right]
=\mathbb{E}\left[\frac{\mathrm{FED}-1}{T}\right].
\label{eq:auc-pec}
\end{equation}
The empirical estimate is
\begin{equation}
\widehat{\mathrm{AUC\mbox{-}PEC}}=\frac{1}{N}\sum_{i=1}^{N}\frac{\mathrm{FED}_i-1}{T_i}.
\label{eq:auc-pec-estimator}
\end{equation}
The second equality in Eq.~\eqref{eq:auc-pec} follows because $\sum_{k=1}^{T}\mathbf{1}\{\mathrm{FED}>k\}=\mathrm{FED}-1$ under the convention $\mathrm{FED}=T+1$ for fully executable trajectories. AUC-PEC lies in $[0,1]$, equals $1$ for fully executable trajectories, and rewards later first failures more than earlier ones. Table headers later omit hats for readability; captions state that reported entries are empirical estimates.

\begin{revision}
The metric family satisfies $0\leq\mathrm{TER}\leq\mathrm{AUC\mbox{-}PEC}\leq1$ in population and sample, and $n_{\mathrm{RR\text{-}Eligible}}=N(1-\widehat{\mathrm{TER}})$ exactly. A fully executable trajectory contributes one to both TER and normalized area; a failing trajectory contributes zero to TER and less than one to area. Let $q=\mathrm{TER}$, $s=\Pr(Y=1\mid\mathrm{FED}=T+1)$, and $r=\mathrm{RR}$. For $0<q<1$,
\begin{equation}
\Pr(Y=1)=q s+(1-q)r.
\label{eq:success-decomposition}
\end{equation}
This partitions terminal success into uninterrupted execution and recovery, without requiring a monotone relationship with TER. Appendix~\ref{app:theory} proves these properties and gives consistency and concentration results under i.i.d. sampling, explicitly distinguished from the finite stratified cohorts studied here.
\end{revision}

\subsection{What Prefix-Level Metrics Add}
\label{sec:metric-value}
\begin{revision}
Table~\ref{tab:metric_validation} summarizes the evaluation signals used in our study and how the proposed prefix-level metrics complement standard outcome and reference-match scores. Exact-call and trajectory-match metrics measure agreement with a reference solution, but they do not localize the first invalid prefix and may penalize alternative executable trajectories. Final success captures task utility, but does not indicate whether the rollout remained executable throughout or succeeded only after an earlier invalid action. PEC, FED, TER, AUC-PEC, and RR fill this gap by measuring prefix survival, first-error depth, full-trajectory executability, normalized executable-prefix area, and recovery after prefix failure.
\end{revision}

\begin{table}[!htbp]
\centering
\small
\begin{tabular}{lcccc}
\toprule
Metric / signal & Ref.-free & Exec.-state & Prefix signal & Recovery \\
\midrule
Exact Call Match & No & No & No & No \\
Trajectory Match & No & No & No & No \\
Final Success ($Y$) & Yes & Yes & No & Indirect \\
$\mathrm{PEC}(k)$ & Yes & Yes & Survival at $k$ & No \\
$\widehat{\mathrm{PEC}}_{\mathrm{elig}}(k)$ & Yes & Yes & Conditional at $k$ & No \\
TER & Yes & Yes & Full horizon & No \\
Mean $\mathrm{FED}_i$ & Yes & Yes & First invalid & No \\
AUC-PEC & Yes & Yes & Normalized profile & No \\
RR & Yes & Yes & Conditional & Yes \\
\bottomrule
\end{tabular}
\caption{Coverage of standard and prefix-level evaluation metrics. Ref.-free means that a metric can credit alternative executable trajectories rather than requiring agreement with a single reference path. Exec.-state means that the metric depends on environment execution or explicit validation semantics. $\mathrm{PEC}(k)$ is the fixed-cohort survival curve in Eq.~\eqref{eq:pec}; $\widehat{\mathrm{PEC}}_{\mathrm{elig}}(k)$ is the empirical eligible-depth diagnostic in Eq.~\eqref{eq:eligible-pec}. TER is the endpoint event $\mathrm{FED}=T+1$, mean $\mathrm{FED}_i$ summarizes first-error depth, AUC-PEC summarizes the horizon-normalized prefix-survival profile, and RR conditions on $\mathrm{FED}\leq T$.}
\label{tab:metric_validation}
\end{table}

\FloatBarrier

\subsection{First-Failure Annotation Protocol}
\label{sec:first-failure-annotation}
When $\mathrm{FED}_i\leq T_i$, we annotate the action at the first invalid prefix with one primary cause: schema failure, dependency failure, precondition failure, state-consistency failure, or information failure. The labels are mutually exclusive by deterministic tie-breaking in that order, so category counts sum to the RR-Eligible set. These labels are diagnostic only; they are not used to compute PEC, TER, AUC-PEC, or RR. The taxonomy is benchmark-agnostic but reflects stressors emphasized by BFCL, ToolSandbox, NESTFUL, FAIL-TaLMs, and function-calling robustness studies~\citep{bfcl, toolsandbox, nestful, fail_talms, robustness_fc}.

\subsection{Validation Intervention}
\label{sec:validation-intervention}
Dual-Loop is included as a controlled validation intervention, not as the main architectural claim. Given the user goal, tool schema, and a compact state summary, a planner proposes a structured plan. A verifier checks tool existence, schema conformity, explicit preconditions, and variable bindings. If verification fails, the planner revises the plan using verifier feedback; once verification succeeds, the approved plan conditions a ReAct executor, which alone issues environment tool calls. Planner and verifier messages are therefore method-side computation, not benchmark tool actions, and all prefix metrics are computed on the executed tool trajectory. The intervention follows planner-executor and verifier-guided generation work, but here serves as a stress test for the evaluation protocol~\citep{adapt, rewoo, plan_and_act, verifier_agents, verifyllm}. \rev{Appendix~\ref{app:dual-loop}} gives the schematic.

\subsection{Benchmarks and Experimental Setup}
\label{sec:benchmarks-setup}

\rev{We use BFCL v3 and ToolSandbox to test} multi-step function-calling correctness and stateful on-policy interaction. BFCL v3 provides multi-turn function-calling scenarios under a standardized interface, making it a test bed for schema, ordering, and cross-step dependency errors~\citep{bfcl}. ToolSandbox adds stateful tool execution, implicit state dependencies, an on-policy user simulator, and dynamic intermediate/final milestone evaluation, making recovery after intermediate failure observable during the rollout~\citep{toolsandbox}.

For each benchmark, we evaluate a stratified sample of $N=200$ instances. The BFCL v3 subset is stratified uniformly over the four multi-turn categories used in this study: \texttt{multi\_turn\_base}, \texttt{multi\_turn\_miss\_func}, \texttt{multi\_turn\_miss\_param}, and \texttt{multi\_turn\_long\_context}, with $50$ tasks from each. The ToolSandbox subset contains $200$ supported open-source scenario templates selected by a deterministic greedy multi-label stratifier over scenario-category labels. Sampling is performed once per dataset, and the same task IDs are reused for every model and method, so all method comparisons are paired at the task level. \rev{The BFCL-v3/ToolSandbox grid comprises} $200 \times 2 \times 3 \times 4=4{,}800$ evaluated rollouts across two datasets, three models, and four agent variants. \rev{We report 95\% task-level confidence intervals. Marginal-interval overlap does not establish paired significance, and we do not infer paired effects from it.} Compute resources for the hosted-inference and local-orchestration components of the \rev{BFCL-v3/ToolSandbox grid} are reported in \rev{Appendix~\ref{app:compute-resources}}.

Both benchmarks are instrumented at the action level. For each rollout, we log whether each predicted call is executable, the first failing step, its primary error label, and whether the episode later reaches benchmark-defined task success. On BFCL v3, this requires prefix-by-prefix replay of the predicted sequence rather than scoring only the terminal trace. On ToolSandbox, we derive the same quantities from executed trajectories and benchmark-provided traces, so state transitions and downstream recovery are observed directly rather than inferred post hoc. \begin{revision}
We evaluate all 50 Airline and 114 Retail base-split tasks in $\tau^3$-bench under the text half-duplex protocol (recorded version v1.0.1). This is complete coverage of those two base splits, not all benchmark domains. One rollout per task, model, and method yields 1,968 Airline/Retail episodes. Across the three benchmarks, the main grid contains 6,768 rollouts over 564 tasks per model--method condition. Appendix~\ref{app:transfer} reports the full scalar metrics, uncertainty, costs, and a separate paired verifier-sensitivity study. Task-level intervals describe variation across tasks under the fixed protocol, not repeated-run variability within a task.
\end{revision}

\subsection{Agent Variants and Experimental Controls}

\rev{We compare a nested intervention ladder with four variants:}
\begin{enumerate}[leftmargin=*, nosep]
    \item \textbf{ReAct}: interleaved reasoning and acting without explicit plan validation~\citep{react}.
    \item \textbf{Prompt Planning}: a planner-first variant in which a plan is generated but not enforced during execution.
    \item \textbf{Plan-Conditioned ReAct}: the generated plan is provided as an explicit control signal during execution.
    \item \textbf{Dual-Loop}: the same plan-conditioned executor augmented with pre-execution verification.
\end{enumerate}
Within each LLM condition, executor calls share the same model setting, tool specification format, decoding strategy, and benchmark horizon cap. Planner and verifier calls occur before execution and are counted as method cost rather than tool actions in $T_i$. Additional provider-specific decoding and rollout settings are reported in \rev{Appendix~\ref{app:compute-resources}}. We report three LLM conditions: GPT-5.4 (\texttt{reasoning.effort=high}), GPT-4.1, and Llama-4-Maverick-Instruct.

\section{Benchmark Results}
\label{sec:benchmarks}

\subsection{Main Benchmark Results}
\rev{Table~\ref{tab:main_results}} reports the main comparison across three LLM conditions using the stratified samples described in \rev{Section~\ref{sec:benchmarks-setup}}. Bracketed intervals are $95\%$ confidence intervals. For TER, mean FED, AUC-PEC, final success, and PEC checkpoints we use $10{,}000$ task-level percentile bootstrap resamples; each resample preserves the shared task IDs across all model-method rollouts within a dataset. RR is a conditional proportion, so its interval is computed with a Wilson score interval using the observed RR-Eligible denominator.

\begin{table}[!htbp]
\centering
\scriptsize
\setlength{\tabcolsep}{3.2pt}
\begin{tabular}{@{}lllcccccc@{}}
\toprule
Dataset & LLM & Method & TER & \shortstack[c]{Mean\\FED} & AUC-PEC & \shortstack[c]{RR-\\Eligible} & RR & \shortstack[c]{Final\\Success} \\
\midrule
BFCL & GPT-5.4 (high) & ReAct & \shortstack[c]{0.69\\{\scriptsize[0.63,0.76]}} & \shortstack[c]{5.61\\{\scriptsize[5.28,5.92]}} & \shortstack[c]{0.83\\{\scriptsize[0.79,0.87]}} & 61 & \shortstack[c]{0.00\\{\scriptsize[0.00,0.06]}} & \shortstack[c]{0.61\\{\scriptsize[0.55,0.68]}} \\
 &  & Prompt Planning & \shortstack[c]{0.78\\{\scriptsize[0.72,0.83]}} & \shortstack[c]{5.72\\{\scriptsize[5.39,6.05]}} & \shortstack[c]{0.85\\{\scriptsize[0.81,0.89]}} & 44 & \shortstack[c]{0.00\\{\scriptsize[0.00,0.08]}} & \shortstack[c]{0.78\\{\scriptsize[0.72,0.83]}} \\
 &  & Plan-Cond. ReAct & \shortstack[c]{0.47\\{\scriptsize[0.40,0.54]}} & \shortstack[c]{4.11\\{\scriptsize[3.73,4.50]}} & \shortstack[c]{0.54\\{\scriptsize[0.48,0.60]}} & 106 & \shortstack[c]{\textbf{0.88}\\{\scriptsize[0.80,0.93]}} & \shortstack[c]{\textbf{0.94}\\{\scriptsize[0.90,0.96]}} \\
 &  & Dual-Loop & \shortstack[c]{\textbf{0.94}\\{\scriptsize[0.91,0.97]}} & \shortstack[c]{\textbf{6.41}\\{\scriptsize[6.12,6.70]}} & \shortstack[c]{\textbf{0.96}\\{\scriptsize[0.93,0.98]}} & 12 & \shortstack[c]{0.00\\{\scriptsize[0.00,0.24]}} & \shortstack[c]{\textbf{0.94}\\{\scriptsize[0.90,0.97]}} \\
 & GPT-4.1 & ReAct & \shortstack[c]{0.62\\{\scriptsize[0.55,0.69]}} & \shortstack[c]{5.18\\{\scriptsize[4.84,5.51]}} & \shortstack[c]{0.76\\{\scriptsize[0.72,0.81]}} & 76 & \shortstack[c]{0.00\\{\scriptsize[0.00,0.05]}} & \shortstack[c]{0.52\\{\scriptsize[0.45,0.58]}} \\
 &  & Prompt Planning & \shortstack[c]{0.79\\{\scriptsize[0.72,0.84]}} & \shortstack[c]{5.84\\{\scriptsize[5.54,6.16]}} & \shortstack[c]{0.87\\{\scriptsize[0.83,0.91]}} & 43 & \shortstack[c]{0.02\\{\scriptsize[0.00,0.12]}} & \shortstack[c]{0.77\\{\scriptsize[0.71,0.82]}} \\
 &  & Plan-Cond. ReAct & \shortstack[c]{0.35\\{\scriptsize[0.28,0.41]}} & \shortstack[c]{3.55\\{\scriptsize[3.16,3.96]}} & \shortstack[c]{0.42\\{\scriptsize[0.35,0.48]}} & 130 & \shortstack[c]{\textbf{0.90}\\{\scriptsize[0.84,0.94]}} & \shortstack[c]{0.93\\{\scriptsize[0.90,0.96]}} \\
 &  & Dual-Loop & \shortstack[c]{\textbf{0.96}\\{\scriptsize[0.94,0.99]}} & \shortstack[c]{\textbf{6.46}\\{\scriptsize[6.18,6.74]}} & \shortstack[c]{\textbf{0.98}\\{\scriptsize[0.95,0.99]}} & 7 & \shortstack[c]{0.00\\{\scriptsize[0.00,0.35]}} & \shortstack[c]{\textbf{0.96}\\{\scriptsize[0.93,0.98]}} \\
 & Llama-4 Maverick & ReAct & \shortstack[c]{0.37\\{\scriptsize[0.30,0.44]}} & \shortstack[c]{4.39\\{\scriptsize[4.07,4.72]}} & \shortstack[c]{0.62\\{\scriptsize[0.57,0.67]}} & 126 & \shortstack[c]{0.01\\{\scriptsize[0.00,0.04]}} & \shortstack[c]{0.27\\{\scriptsize[0.20,0.33]}} \\
 &  & Prompt Planning & \shortstack[c]{0.60\\{\scriptsize[0.54,0.67]}} & \shortstack[c]{5.51\\{\scriptsize[5.17,5.84]}} & \shortstack[c]{0.79\\{\scriptsize[0.75,0.83]}} & 79 & \shortstack[c]{0.00\\{\scriptsize[0.00,0.05]}} & \shortstack[c]{0.56\\{\scriptsize[0.48,0.62]}} \\
 &  & Plan-Cond. ReAct & \shortstack[c]{0.05\\{\scriptsize[0.02,0.08]}} & \shortstack[c]{1.96\\{\scriptsize[1.78,2.13]}} & \shortstack[c]{0.15\\{\scriptsize[0.12,0.18]}} & 190 & \shortstack[c]{\textbf{0.73}\\{\scriptsize[0.66,0.79]}} & \shortstack[c]{0.74\\{\scriptsize[0.68,0.80]}} \\
 &  & Dual-Loop & \shortstack[c]{\textbf{0.88}\\{\scriptsize[0.83,0.92]}} & \shortstack[c]{\textbf{6.22}\\{\scriptsize[5.92,6.53]}} & \shortstack[c]{\textbf{0.93}\\{\scriptsize[0.89,0.95]}} & 25 & \shortstack[c]{0.00\\{\scriptsize[0.00,0.13]}} & \shortstack[c]{\textbf{0.79}\\{\scriptsize[0.72,0.84]}} \\
\midrule
ToolSandbox & GPT-5.4 (high) & ReAct & \shortstack[c]{0.74\\{\scriptsize[0.68,0.80]}} & \shortstack[c]{3.84\\{\scriptsize[3.63,4.04]}} & \shortstack[c]{0.83\\{\scriptsize[0.78,0.87]}} & 52 & \shortstack[c]{0.27\\{\scriptsize[0.17,0.40]}} & \shortstack[c]{0.59\\{\scriptsize[0.52,0.66]}} \\
 &  & Prompt Planning & \shortstack[c]{0.82\\{\scriptsize[0.77,0.88]}} & \shortstack[c]{4.08\\{\scriptsize[3.89,4.26]}} & \shortstack[c]{0.88\\{\scriptsize[0.84,0.92]}} & 35 & \shortstack[c]{0.17\\{\scriptsize[0.08,0.33]}} & \shortstack[c]{0.61\\{\scriptsize[0.54,0.68]}} \\
 &  & Plan-Cond. ReAct & \shortstack[c]{0.48\\{\scriptsize[0.41,0.56]}} & \shortstack[c]{2.88\\{\scriptsize[2.65,3.13]}} & \shortstack[c]{0.52\\{\scriptsize[0.46,0.59]}} & 103 & \shortstack[c]{\textbf{0.73}\\{\scriptsize[0.64,0.80]}} & \shortstack[c]{\textbf{0.69}\\{\scriptsize[0.62,0.76]}} \\
 &  & Dual-Loop & \shortstack[c]{\textbf{0.94}\\{\scriptsize[0.90,0.96]}} & \shortstack[c]{\textbf{4.32}\\{\scriptsize[4.15,4.47]}} & \shortstack[c]{\textbf{0.95}\\{\scriptsize[0.93,0.98]}} & 13 & \shortstack[c]{0.15\\{\scriptsize[0.04,0.42]}} & \shortstack[c]{0.63\\{\scriptsize[0.56,0.69]}} \\
 & GPT-4.1 & ReAct & \shortstack[c]{0.72\\{\scriptsize[0.66,0.78]}} & \shortstack[c]{3.84\\{\scriptsize[3.64,4.03]}} & \shortstack[c]{0.84\\{\scriptsize[0.80,0.88]}} & 56 & \shortstack[c]{0.20\\{\scriptsize[0.11,0.32]}} & \shortstack[c]{0.49\\{\scriptsize[0.42,0.56]}} \\
 &  & Prompt Planning & \shortstack[c]{0.81\\{\scriptsize[0.76,0.86]}} & \shortstack[c]{4.11\\{\scriptsize[3.93,4.29]}} & \shortstack[c]{0.89\\{\scriptsize[0.85,0.93]}} & 37 & \shortstack[c]{0.22\\{\scriptsize[0.11,0.37]}} & \shortstack[c]{0.56\\{\scriptsize[0.49,0.63]}} \\
 &  & Plan-Cond. ReAct & \shortstack[c]{0.35\\{\scriptsize[0.29,0.42]}} & \shortstack[c]{2.45\\{\scriptsize[2.22,2.69]}} & \shortstack[c]{0.40\\{\scriptsize[0.34,0.46]}} & 129 & \shortstack[c]{\textbf{0.59}\\{\scriptsize[0.50,0.67]}} & \shortstack[c]{\textbf{0.62}\\{\scriptsize[0.56,0.69]}} \\
 &  & Dual-Loop & \shortstack[c]{\textbf{0.96}\\{\scriptsize[0.93,0.98]}} & \shortstack[c]{\textbf{4.38}\\{\scriptsize[4.22,4.53]}} & \shortstack[c]{\textbf{0.97}\\{\scriptsize[0.95,0.99]}} & 8 & \shortstack[c]{0.38\\{\scriptsize[0.14,0.69]}} & \shortstack[c]{\textbf{0.62}\\{\scriptsize[0.56,0.69]}} \\
 & Llama-4 Maverick & ReAct & \shortstack[c]{0.49\\{\scriptsize[0.42,0.56]}} & \shortstack[c]{3.33\\{\scriptsize[3.12,3.54]}} & \shortstack[c]{0.68\\{\scriptsize[0.63,0.74]}} & 101 & \shortstack[c]{0.23\\{\scriptsize[0.16,0.32]}} & \shortstack[c]{0.38\\{\scriptsize[0.31,0.44]}} \\
 &  & Prompt Planning & \shortstack[c]{0.64\\{\scriptsize[0.57,0.70]}} & \shortstack[c]{3.69\\{\scriptsize[3.50,3.89]}} & \shortstack[c]{0.78\\{\scriptsize[0.73,0.82]}} & 73 & \shortstack[c]{0.18\\{\scriptsize[0.11,0.28]}} & \shortstack[c]{0.49\\{\scriptsize[0.42,0.56]}} \\
 &  & Plan-Cond. ReAct & \shortstack[c]{0.09\\{\scriptsize[0.06,0.13]}} & \shortstack[c]{1.63\\{\scriptsize[1.49,1.79]}} & \shortstack[c]{0.15\\{\scriptsize[0.11,0.19]}} & 182 & \shortstack[c]{\textbf{0.57}\\{\scriptsize[0.50,0.64]}} & \shortstack[c]{0.56\\{\scriptsize[0.48,0.62]}} \\
 &  & Dual-Loop & \shortstack[c]{\textbf{0.80}\\{\scriptsize[0.74,0.85]}} & \shortstack[c]{\textbf{4.13}\\{\scriptsize[3.96,4.30]}} & \shortstack[c]{\textbf{0.90}\\{\scriptsize[0.87,0.93]}} & 40 & \shortstack[c]{0.35\\{\scriptsize[0.22,0.50]}} & \shortstack[c]{\textbf{0.61}\\{\scriptsize[0.54,0.68]}} \\
\bottomrule
\end{tabular}

\caption{Benchmark results with $200$ stratified sampled instances for BFCL multi-turn and $200$ multi-label-stratified sampled instances for ToolSandbox under three LLM conditions. All entries are empirical estimates; table headers omit hats for readability. Bracketed values are $95\%$ confidence intervals. Intervals for TER, mean FED, AUC-PEC, and final success use $10{,}000$ task-level percentile bootstrap resamples; RR intervals use Wilson score intervals with the RR-Eligible denominator. RR-Eligible counts episodes with $\mathrm{FED}\leq T$ and is reported without an interval. \rev{Bold point estimates mark column maxima within each dataset--LLM block, including ties at the displayed precision; boldface does not indicate statistical significance.} \rev{Higher TER, AUC-PEC, and final success indicate stronger respective scores; mean FED must also be interpreted with rollout horizon;} RR is better only conditional on entering the failure regime and should be interpreted with RR-Eligible.}
\label{tab:main_results}
\end{table}

\rev{On BFCL-v3 and ToolSandbox, the point estimates show a consistent separation between terminal success and prefix reliability.} Under GPT-5.4, Plan-Cond. ReAct reaches the highest or tied-highest final success on BFCL and ToolSandbox ($0.94$ and $0.69$), but its prefix reliability is much weaker than Dual-Loop's: TER is $0.47$ and $0.48$, and AUC-PEC is $0.54$ and $0.52$. Dual-Loop instead attains TER of $0.94$ on both datasets and AUC-PEC of $0.96$ and $0.95$. The same gap remains under weaker LLMs: Dual-Loop exceeds Plan-Cond. ReAct in TER by $0.61$ under GPT-4.1 and by $0.83$ under Llama-4 Maverick on BFCL, and by $0.61$ and $0.71$ on ToolSandbox.

AUC-PEC and mean FED clarify how outcome-only and prefix-level rankings separate. On ToolSandbox with GPT-5.4, Plan-Cond. ReAct has the highest final success ($0.69$), but Dual-Loop has much higher TER, mean FED, and AUC-PEC ($0.94$, $4.32$, and $0.95$ versus $0.48$, $2.88$, and $0.52$). 

\begin{revision}
\paragraph{Success after a prefix failure.}\label{sec:recovery-analysis}
The BFCL-v3/Llama-4 Maverick Plan-Conditioned ReAct condition makes Eq.~\eqref{eq:success-decomposition} concrete. Of 200 episodes, 10 were fully executable and all 10 succeeded; 138 of the 190 episodes with a prefix failure also succeeded. Thus TER is $10/200=0.05$, RR is $138/190\approx0.7263$, and final success is $(10+138)/200=0.74$. Fully $138/148=93.2\%$ of successful episodes reached the objective after an earlier invalid prefix. This is high terminal utility with low uninterrupted-execution reliability, rather than a contradiction between metrics. Appendix~\ref{app:contact-repair} illustrates how a rejected contact-removal attempt can be repaired by first retrieving the required identifier.
\end{revision}

RR should be read together with RR-Eligible. Across the three LLMs, Dual-Loop enters the RR-eligible regime far less often than Plan-Cond. ReAct: on BFCL the counts are $12$, $7$, and $25$ for Dual-Loop versus $106$, $130$, and $190$ for Plan-Cond. ReAct, and on ToolSandbox they are $13$, $8$, and $40$ versus $103$, $129$, and $182$. \rev{Dual-Loop has zero observed RR on BFCL; the small eligible sets make these conditional estimates imprecise. Plan-Conditioned ReAct combines many more prefix failures with substantial success after those failures, so its RR measures recovery rather than uninterrupted executability. The size of the eligible set does not itself determine the recovery rate.}

\FloatBarrier
\subsection{Prefix Survival by Depth}

\begin{figure}[!htbp]
\centering
\includegraphics[width=0.93\linewidth]{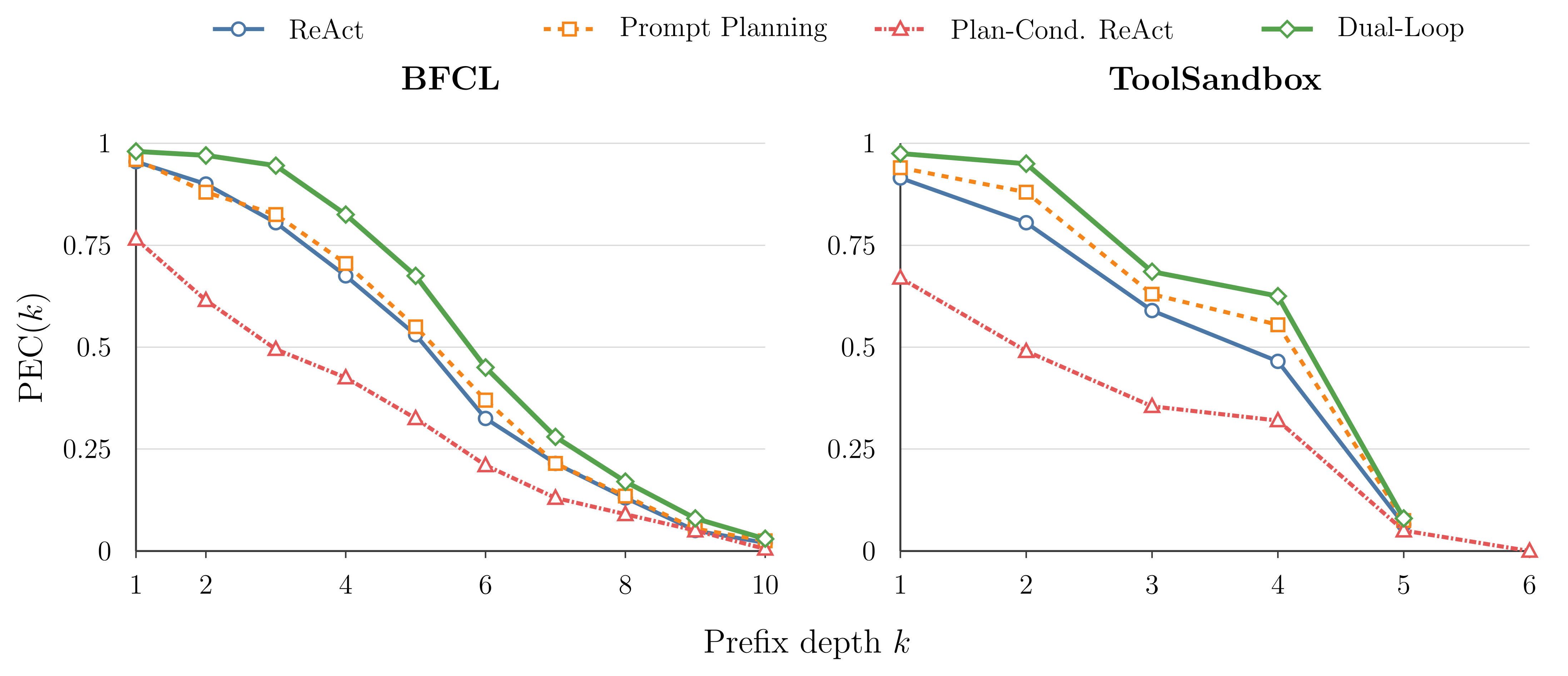}
\caption{Prefix-executability curves on the $200$-instance BFCL multi-turn sample and the $200$-instance ToolSandbox sample using GPT-5.4 (high). Each point plots $\widehat{\mathrm{PEC}}(k)$ with the fixed-cohort estimator in Eq.~\eqref{eq:pec-estimator}. \rev{The BFCL panel displays depths $k\leq 10$; under Eq.~\eqref{eq:pec}, late-depth declines can reflect both prefix failures and termination.} \rev{Appendix Table~\ref{tab:pec_depths}} reports early-depth checkpoints for all LLM conditions.}
\label{fig:pec_curves}
\end{figure}

Figure~\ref{fig:pec_curves} shows the fixed-cohort PEC curves for GPT-5.4; \rev{Appendix Table~\ref{tab:pec_depths}} provides the corresponding checkpoints for all three LLM conditions. \rev{At early depths, the BFCL-v3/ToolSandbox audit shows large fixed-cohort differences. For example, BFCL PEC@4 for Plan-Conditioned ReAct is $0.42$, $0.33$, and $0.05$ across the three model conditions, versus $0.82$, $0.82$, and $0.78$ for Dual-Loop. Under Eq.~\eqref{eq:pec-support}, these are joint reach-and-validity statistics. Consequently, a fixed-cohort decline need not represent a newly invalid call; depth support and conditional validity answer complementary questions.}

\FloatBarrier

\begin{revision}
\subsection{Transfer, Cost, and the Boundary of Validation}
The $\tau^3$-bench audit again separates terminal outcome from prefix reliability (Table~\ref{tab:tau-results}). With GPT-5.4, Plan-Conditioned ReAct has the highest final-success point estimate ($0.51$), while ReAct has the highest TER/AUC-PEC estimates ($0.95/0.95$). With GPT-4.1, Prompt Planning leads final success ($0.51$), while ReAct leads TER/AUC-PEC ($0.84/0.91$). With Llama-4, ReAct, Prompt Planning, and Plan-Conditioned ReAct tie at the displayed final-success precision ($0.18$), despite TER spanning $0.52$--$0.60$ and AUC-PEC $0.56$--$0.65$. These are model-dependent point-estimate orderings, not claims of paired significance.

The Airline/Retail audit does not establish the consistent validation advantage observed on BFCL-v3 and ToolSandbox. A possible explanation is that BFCL emphasizes pre-specified composition, whereas interactive user feedback and state changes in $\tau^3$-bench can make advance plans stale. This is a testable interpretation, not an identified causal mechanism.

Measured agent-side costs make the trade-off explicit. In the four-method $\tau^3$ audit, Dual-Loop changes mean tokens/cumulative agent-call latency relative to Plan-Conditioned ReAct by approximately $+3\%/+22\%$, $+12\%/+7\%$, and $-3\%/-1\%$ for GPT-5.4, GPT-4.1, and Llama-4, respectively. These totals include verification and revision as well as any downstream execution savings; they exclude simulator and evaluator calls. Table~\ref{tab:tau-cost} gives all four methods and accounting details.

A separate sensitivity study fixes the initial plan and GPT-5.4-high planner/executor across all 164 tasks, varying verifier model and validation budget. All paired 95\% intervals for TER and AUC-PEC changes include zero. The GPT-5.4 verifier at $K_{\mathrm{val}}=5$, for example, changes TER by $+0.006$ $[-0.018,0.030]$ while adding 40.7k tokens and 176.9s relative to its paired Plan-Conditioned rerun. The deterministic-acceptance wrapper also changes executor guidance and is not an equivalent-policy no-verifier control. Table~\ref{tab:verifier-sensitivity} reports the full comparisons and limitations. These findings delimit the benefit of validation in this environment; they do not prove equivalence or isolate verification logic from extra computation.
\end{revision}

\FloatBarrier

\section{Conclusion}

Reliable tool use is not only a question of whether an agent eventually succeeds, but whether its trajectory remains executable as it unfolds. This paper introduced prefix executability as an evaluation protocol for tool-using agents, formalizing PEC, FED, TER, AUC-PEC, and RR for variable-horizon rollouts. \rev{Across the three-benchmark audit, prefix reliability and terminal outcome provide complementary information. Validation improves prefix reliability on BFCL-v3 and ToolSandbox, while the Airline/Retail transfer audit and paired sensitivity study show that this benefit is environment-dependent. Exact identities explain how high final success can coexist with frequent prefix failure, and cost measurements expose the operational price of validation. Evaluations should therefore report uninterrupted executability, recovery, terminal success, and agent-side cost together.}

Although we report PEC, FED, TER, AUC-PEC, and RR as retrospective evaluation
metrics over completed rollouts, the underlying executable-prefix predicate is
prefix-local. When available during execution, it can be checked after each
candidate tool action to support runtime monitoring, repair, replanning, or
human escalation before invalid state is propagated. Aggregated over rollouts,
the same signals can guide offline choices such as prompt design, validation
budgets, and agent selection. These uses remain complementary to the paper's
evaluation focus: prefix executability exposes whether and where a trajectory
becomes invalid, but it does not by itself certify semantic correctness, safety,
privacy, or fairness. Developing closed-loop agents and deployment policies that
optimize against these signals is an important direction for future work.

\section*{Limitations}
\begin{revision}
The study is an audit rather than a complete leaderboard evaluation: BFCL-v3 and ToolSandbox each contribute 200 sampled tasks, ToolSandbox is restricted to supported local-tool scenarios, and $\tau^3$ coverage is complete only for the Airline/Retail base splits. Task-level intervals do not measure within-task repeated-seed variation. The main-grid tables provide marginal intervals; paired intervals are reported for the separate verifier-sensitivity study. Fixed-cohort PEC reflects both reaching a depth and retaining validity, so its decline is not an error-only statistic.

Executability assumes explicit schemas, preconditions, and benchmark-accessible state. Partially observed web or GUI environments may require state estimation or learned preconditions, and executability does not certify semantic correctness. We did not evaluate a matched schema-constrained-decoding baseline, so the reported rankings are not established under such constraints. These constraints can prevent malformed arguments, while prefix checking additionally covers object existence, provenance, state preconditions, and transition acceptance.

The intervention combines validation/revision with extra inference. The sensitivity study varies verifier capacity and budget, but its acceptance wrapper changes executor guidance; it does not fully separate validation logic from compute or context effects. Per-method cost telemetry is available for the $\tau^3$ audit, whereas the BFCL-v3/ToolSandbox grid has aggregate runtime only. Broader claims require additional domains, repeated runs, and matched constrained-decoding and repair controls.
\end{revision}

\paragraph{Broader impacts.}
Prefix-level executability can improve oversight of tool-using agents by
identifying invalid API calls, unresolved dependencies, and state-precondition
violations before they are obscured by final-success scores. This is most useful
in workflows where tool calls modify persistent state, consume resources,
trigger external actions, or require human escalation. The same diagnostics may
also make autonomous tool use easier to deploy in settings with privacy,
security, fairness, or service-access risks. Moreover, an executable trajectory
can still be unsafe, biased, privacy-violating, or misaligned with user intent.
Prefix executability should therefore be treated as one reliability signal and
combined with domain-specific safety policies, privacy controls, security
review, and human oversight in consequential applications.

\section*{Acknowledgments}
\rev{This work was funded by Oracle.}

\begingroup
\small
\urlstyle{same}
\bibliography{custom}
\endgroup

\clearpage
\appendix
\makeatletter
\@addtoreset{figure}{section}
\@addtoreset{table}{section}
\makeatother
\renewcommand{\thefigure}{\thesection.\arabic{figure}}
\renewcommand{\thetable}{\thesection.\arabic{table}}

\begin{revision}
\section{Relationship to Prior Evaluation Protocols}
\label{app:protocol-comparison}

Table~\ref{tab:trajectory-comparison} compares the evaluation units and signals of related protocols. Prefix executability measures whether the observed action sequence satisfies the benchmark's schemas, state preconditions, and data dependencies. Alternative action sequences receive full executability credit whenever these conditions hold, regardless of agreement with a reference trajectory. Prefix metrics therefore complement evaluations of task outcomes, intermediate milestones, reference-trajectory agreement, and semantic process quality.

\begin{table}[!htbp]
\centering
\small
\setlength{\tabcolsep}{4pt}
\renewcommand{\arraystretch}{1.14}
\begin{tabularx}{\linewidth}{@{}>{\raggedright\arraybackslash}p{0.21\linewidth}>{\raggedright\arraybackslash}X>{\raggedright\arraybackslash}X@{}}
\toprule
Evaluation & Primary signal & Relation to prefix executability \\
\midrule
T-Eval~\citep{teval}
& Component capabilities: planning, reasoning, retrieval, understanding, instruction following, and review.
& Diagnoses component abilities; PEC tracks executable prefixes across depths, and FED identifies the first invalid action. \\
ToolSandbox~\citep{toolsandbox}
& Task-specified required and prohibited conditions over states and traces.
& Checks required or prohibited conditions; PEC/FED localize first non-executability and survival over depth. \\
$\tau$-bench~\citep{taubench} and $\tau^3$-bench~\citep{tauthree}
& Task outcomes in policy-constrained user--agent--environment interaction, including terminal goal-state checks.
& Provide stateful environments where prefix metrics complement terminal reward. \\
TRAJECT-Bench~\citep{traject_bench}
& Reference-based tool/argument/order matching and reference-free, LLM-judged trajectory satisfaction.
& Combines reference-based and judged trajectory quality; prefix metrics measure benchmark-grounded executability, first failure, and recovery. \\
ToolPRMBench~\citep{toolprmbench}
& Correct/incorrect step cases for process-reward-model evaluation.
& Assesses process judges; prefix metrics use benchmark execution semantics without a learned reward model. \\
AgentProcessBench\newline\citep{agentprocessbench}
& Human-labeled step effectiveness; step and first-error prediction accuracy.
& Evaluates agreement with human process labels; prefix metrics directly test benchmark execution feasibility. \\
TRACE~\citep{trace}
& Reference-free LLM assessment of efficiency, hallucination, and adaptivity using an evidence bank.
& Assesses broader trajectory quality; prefix metrics add deterministic feasibility, first failure, and recovery. \\
Procedure-Aware Evaluation~\citep{procedure_aware}
& Utility, efficiency, interaction quality, and procedural integrity.
& Audits broader procedural quality; prefix metrics isolate execution feasibility. \\
Beyond Success Rate~\citep{beyond_success_rate}
& Recovery under deliberately injected tool/environment faults.
& Measures recovery per injected fault; RR here measures terminal success conditional on prefix failure observed without injected faults. \\
\bottomrule
\end{tabularx}
\caption{Evaluation units and signals across trajectory and process evaluation protocols.}
\label{tab:trajectory-comparison}
\end{table}

\FloatBarrier
\end{revision}

\FloatBarrier
\begin{revision}
\section{Exact Properties, Consistency, and Relationship to Final Success}
\label{app:theory}

The properties below distinguish identities that hold for every evaluated cohort from population guarantees that require a sampling assumption. Throughout, rollout $i$ has an evaluated horizon $T_i\geq 1$, first-error depth $F_i:=\mathrm{FED}_i\in\{1,\ldots,T_i+1\}$, and final-success indicator $Y_i\in\{0,1\}$. The value $F_i=T_i+1$ denotes a fully executable trajectory. Let $(T,F,Y)$ denote the corresponding population variables. By definition, an executable prefix contains only executable shorter prefixes. No independence assumption is needed for Propositions~1--3.

\paragraph{Proposition 1 (fixed-cohort monotonicity).}
For an integer depth $k\geq 1$, define
\begin{equation}
E_{i,k}:=\mathbf{1}\{T_i\geq k\ \land\ \operatorname{Exec}(\hat{\tau}_{i,\leq k})\}
=\mathbf{1}\{F_i>k\},\qquad
\widehat{\mathrm{PEC}}(k)=\frac{1}{N}\sum_{i=1}^{N}E_{i,k}.
\label{eq:theory-prefix-indicator}
\end{equation}
Then both the empirical curve and its population counterpart $\mathrm{PEC}(k):=\Pr(F>k)$ are non-increasing:
\begin{equation}
1\geq\widehat{\mathrm{PEC}}(k)\geq\widehat{\mathrm{PEC}}(k+1)\geq0,
\qquad
1\geq\mathrm{PEC}(k)\geq\mathrm{PEC}(k+1)\geq0.
\label{eq:theory-monotonicity}
\end{equation}
\emph{Proof.} If $E_{i,k+1}=1$, rollout $i$ reaches depth $k+1$ with an executable prefix. It therefore reaches depth $k$ with an executable prefix, so $E_{i,k}=1$. Thus $E_{i,k+1}\leq E_{i,k}$ pointwise; averaging or taking expectations proves the result. The denominator remains $N$ at every depth. A rollout ending before depth $k$ contributes zero to the fixed-cohort event in Eq.~\eqref{eq:theory-prefix-indicator}; the statement does not require a censoring model.

\paragraph{Proposition 2 (bounds and finite-cohort identities).}
Define $G_i:=\mathbf{1}\{F_i=T_i+1\}$ and $A_i:=(F_i-1)/T_i$. Then
\begin{align}
0&\leq\widehat{\mathrm{TER}}\leq\widehat{\mathrm{AUC\text{-}PEC}}\leq1,
&0&\leq\mathrm{TER}\leq\mathrm{AUC\text{-}PEC}\leq1,
\label{eq:theory-area-bounds}\\
n_{\mathrm{RR\text{-}Eligible}}&=\sum_{i=1}^{N}\mathbf{1}\{F_i\leq T_i\}
=N\bigl(1-\widehat{\mathrm{TER}}\bigr).
\label{eq:theory-rr-identity}
\end{align}
The empirical mean $\overline{F}:=N^{-1}\sum_iF_i$ satisfies
\begin{equation}
1\leq\overline{F}\leq\frac{1}{N}\sum_{i=1}^{N}(T_i+1).
\label{eq:theory-fed-bounds}
\end{equation}
If $\mathbb{E}[T]<\infty$, its population analogue is $1\leq\mathbb{E}[F]\leq\mathbb{E}[T]+1$. When every rollout has the same horizon $T_0$,
\begin{equation}
\widehat{\mathrm{TER}}=\widehat{\mathrm{PEC}}(T_0),\qquad
\widehat{\mathrm{AUC\text{-}PEC}}=\frac{\overline{F}-1}{T_0}.
\label{eq:theory-fixed-horizon}
\end{equation}
\emph{Proof.} A fully executable trajectory has $G_i=A_i=1$. Otherwise, $G_i=0$ and $0\leq A_i<1$. Hence $0\leq G_i\leq A_i\leq1$ pointwise, proving Eq.~\eqref{eq:theory-area-bounds} by averaging and taking expectations. The events $F_i=T_i+1$ and $F_i\leq T_i$ partition the cohort, giving Eq.~\eqref{eq:theory-rr-identity}. The bound $1\leq F_i\leq T_i+1$ gives Eq.~\eqref{eq:theory-fed-bounds} and its expectation counterpart. Finally, when $T_i=T_0$, the events $F_i>T_0$ and $F_i=T_0+1$ coincide, and the normalization $T_0$ can be taken outside the empirical average.

For variable horizons, TER remains the complete-trajectory event $F=T+1$, whereas AUC-PEC is the mean per-trajectory normalized area $\mathbb{E}[(F-1)/T]$. Thus the common-horizon identities in Eq.~\eqref{eq:theory-fixed-horizon} must not be applied to variable-length rollouts. In particular, mean FED is generally not determined by TER or AUC-PEC alone.

\paragraph{Proposition 3 (final-success decomposition).}
Let $G:=\{F=T+1\}$ and $B:=\{F\leq T\}$ denote the fully executable and prefix-failure regimes. For every rollout distribution,
\begin{equation}
\Pr(Y=1)=\Pr(\{Y=1\}\cap G)+\Pr(\{Y=1\}\cap B).
\label{eq:theory-success-partition}
\end{equation}
If $0<q:=\Pr(G)<1$, define $s:=\Pr(Y=1\mid G)$ and $r:=\Pr(Y=1\mid B)=\mathrm{RR}$. Since $q=\mathrm{TER}$,
\begin{equation}
\Pr(Y=1)=q\,s+(1-q)\,r,\qquad
(1-q)r\leq\Pr(Y=1)\leq q+(1-q)r.
\label{eq:theory-success-decomposition}
\end{equation}
\emph{Proof.} The events $G$ and $B$ are disjoint and exhaustive, proving Eq.~\eqref{eq:theory-success-partition}. When both have positive probability, the definition of conditional probability gives the decomposition in Eq.~\eqref{eq:theory-success-decomposition}; its bounds follow from $0\leq s\leq1$. At $q=0$ or $q=1$, the conditional probability on the absent regime is undefined, and Eq.~\eqref{eq:theory-success-partition} supplies the valid statement without assigning a value to that conditional. The same partition argument applies to empirical counts, with conditional empirical rates defined only for nonempty regimes.

Terminal success therefore combines successful fully executable trajectories and successful trajectories with an earlier prefix break. Holding $s$ and $r$ fixed, its slope with respect to TER is $s-r$; across methods, both conditional success rates can also change. Consequently, the decomposition does not imply a universal monotone relationship between TER and final success. Section~\ref{sec:recovery-analysis} gives a concrete empirical instance of this distinction.

\paragraph{Proposition 4 (consistency and concentration under i.i.d. sampling).}
Assume that task--rollout pairs are independent and identically distributed draws from a fixed evaluation distribution. For a fixed depth $k$, let $Z_i$ be any of
\begin{equation}
E_{i,k},\qquad \mathbf{1}\{F_i=T_i+1\},\qquad
\frac{F_i-1}{T_i},\qquad Y_i,
\label{eq:theory-bounded-variables}
\end{equation}
corresponding to PEC at $k$, TER, AUC-PEC, and final success, respectively. Then $\widehat{\theta}:=N^{-1}\sum_i Z_i$ is strongly consistent for $\theta:=\mathbb{E}[Z_i]$, and, for every $\epsilon>0$,
\begin{equation}
\Pr\bigl(|\widehat{\theta}-\theta|\geq\epsilon\bigr)
\leq2\exp(-2N\epsilon^2).
\label{eq:theory-hoeffding}
\end{equation}
Mean FED is strongly consistent for $\mathbb{E}[F]$ if $\mathbb{E}[T]<\infty$. If, additionally, $T\leq K$ with probability one for a finite constant $K$, then
\begin{equation}
\Pr\bigl(|\overline{F}-\mathbb{E}[F]|\geq\epsilon\bigr)
\leq2\exp\!\left(-\frac{2N\epsilon^2}{K^2}\right).
\label{eq:theory-fed-concentration}
\end{equation}
Finally, if $\Pr(B)>0$, the recovery-rate estimator
\begin{equation}
\widehat{\mathrm{RR}}
=\frac{N^{-1}\sum_{i=1}^{N}Y_i\mathbf{1}\{F_i\leq T_i\}}
{N^{-1}\sum_{i=1}^{N}\mathbf{1}\{F_i\leq T_i\}}
\label{eq:theory-rr-ratio}
\end{equation}
has a nonzero denominator once at least one eligible episode has been sampled. With probability one, such an episode occurs as the sample grows, and $\widehat{\mathrm{RR}}$ converges to the population recovery rate $\Pr(Y=1\mid B)$ as $N\to\infty$.

\emph{Proof.} Every variable in Eq.~\eqref{eq:theory-bounded-variables} lies in $[0,1]$. The strong law of large numbers establishes strong consistency, and Hoeffding's inequality gives Eq.~\eqref{eq:theory-hoeffding}. Since $1\leq F\leq T+1$, the condition $\mathbb{E}[T]<\infty$ ensures that $F$ is integrable, so the strong law also applies to mean FED. Under the bounded-horizon assumption, $F\in[1,K+1]$ has range of width $K$; Hoeffding's inequality yields Eq.~\eqref{eq:theory-fed-concentration}. For RR, the numerator and denominator of Eq.~\eqref{eq:theory-rr-ratio} converge with probability one to $\Pr(\{Y=1\}\cap B)$ and $\Pr(B)$, respectively. When $\Pr(B)>0$, continuity of division at a positive denominator proves the result. For a finite cohort with no eligible episodes, RR remains undefined.

\paragraph{Interpretation for the reported audits.}
The finite-cohort identities and monotonicity above hold exactly for the reported evaluations, including the deterministic stratified cohorts. Population consistency and concentration hold under the stated i.i.d. sampling assumption. The success decomposition separates success on fully executable trajectories from success after a prefix failure, making recovery explicit alongside prefix reliability.
\end{revision}

\FloatBarrier
\begin{revision}
\section{Deterministic Executability Checking and Recovery}
\label{app:checker}

\subsection{Executability Checks and First-Failure Classification}
\label{app:checker-procedure}

The executability scorer is deterministic and separate from the LLM-based pre-execution verifier used by Dual-Loop. The scorer evaluates the actual emitted actions against benchmark tool and state semantics; no learned judge, human step label, or single gold action sequence is required. The Dual-Loop verifier instead examines a proposed plan before execution and may change the executor's subsequent behavior. Its acceptance decision does not establish that the emitted trajectory is executable.

For action $a_j$, let $V_j=1$ exactly when the tool is exposed, its arguments satisfy the declared interface and validators, referenced values are available and type-compatible, required information has been obtained, applicable state preconditions hold, and the benchmark accepts the transition. Otherwise $V_j=0$. The checker determines first failure as follows.
\begin{enumerate}[leftmargin=*, itemsep=3pt]
    \item Initialize replay from the benchmark's initial state. Available values comprise the prompt and benchmark-visible initial values. Initialize $\mathrm{FED}=T+1$.
    \item Process emitted actions in order. Resolve the selected tool; an unavailable tool is a schema/tool-availability failure. For an exposed tool, check argument binding, declared types, and registered validators.
    \item Check referenced values for absence, stale bindings, or incompatible types; these are dependency failures. Check whether required task information has been obtained, and whether applicable state preconditions hold.
    \item If these checks pass, execute the action against the replay or sandbox state. A rejected transition is a state-consistency failure. For an accepted transition, record the output, update available values, and advance the state.
    \item At the first failed predicate or rejected transition, set $V_j=0$ and $\mathrm{FED}=j$. Choose the first applicable diagnostic category in the precedence order below, then stop prefix replay. Retain the full observed rollout for benchmark-defined final success $Y$ and RR.
    \item If every action satisfies the predicate, retain $\mathrm{FED}=T+1$. A terminal no-tool action is executable only when benchmark semantics permit completion without a tool call.
\end{enumerate}
The diagnostic precedence is
\[
\text{schema}\ \longrightarrow\ \text{dependency}\ \longrightarrow\
\text{precondition}\ \longrightarrow\ \text{state-consistency}\
\longrightarrow\ \text{information}.
\]
The tie-break assigns one primary cause at the first failed action; it changes neither action validity nor FED, PEC, TER, AUC-PEC, RR, or final success. In particular, the diagnostic ordering is distinct from the order in which applicable conditions are checked.

BFCL-v3 evaluates prefixes by replay in a copied state under the case-specific function namespace and benchmark evaluator semantics~\citep{bfcl}. ToolSandbox uses tools visible at the relevant turn, their registered validators, and the recorded execution outcome and state transition~\citep{toolsandbox}. These benchmark-specific semantics ground the same executability predicate. The metrics then aggregate the resulting $(T_i,\mathrm{FED}_i,Y_i)$ records deterministically.

Stopping the first-failure replay does not truncate the observed episode or replace its final outcome. Here $V_j$ describes the validity of action $a_j$ in its local benchmark state. Determining FED requires checking actions only through the first failure. Later actions in the observed episode can be executable, and the task can still succeed. For $k\leq T$, however, prefix validity is
\[
E_k=\prod_{j=1}^{k} V_j=\mathbf{1}\{\mathrm{FED}>k\}.
\]
Once an action fails, every subsequent prefix still contains that failure. Consequently, later successful execution can contribute to recovery while never restoring executable-prefix survival for that rollout.

\subsection{Illustrative Contact-Repair Comparison}
\label{app:contact-repair}

Table~\ref{tab:contact-repair} illustrates a contact-removal task under ToolSandbox semantics. In the first action, the agent supplies a phone number in place of the stored contact identifier. Although both values have string type, the requested identifier is absent from the current state, so the action is rejected without changing the state. A subsequent lookup obtains the correct identifier and enables removal.

\begin{table}[!htbp]
\centering
\small
\begin{tabularx}{\linewidth}{@{}p{0.07\linewidth}XX@{}}
\toprule
Step & Emitted action & Benchmark outcome \\
\midrule
1 & Attempt to remove a contact using its phone number as its stored identifier. & Rejected because no matching identifier exists; state unchanged. \\
2 & Search for the contact by phone number. & Returns the contact's stored identifier. \\
3 & Remove the contact using the returned identifier. & Accepted; the requested contact is removed. \\
\bottomrule
\end{tabularx}
\caption{Illustrative repair after an invalid first action. The complete observed trajectory is retained even though prefix survival ends at the first step.}
\label{tab:contact-repair}
\end{table}

Here $V=(0,1,1)$ but $E=(0,0,0)$. For $T=3$, $\mathrm{FED}=1$, the trajectory contributes $0$ to TER and AUC-PEC, and it is RR-eligible; its terminal success can nevertheless be $Y=1$. An error-free alternative first searches for the contact and then removes it using the returned identifier. Both actions in this two-step sequence are executable, so the full trajectory receives executability credit.

A terminal ToolSandbox milestone could credit both the repaired and error-free trajectories. A task-specific check for prohibited actions could distinguish them, while prefix executability directly records the rejected transition. TRAJECT-Bench could penalize the extra action or recognize the later dependency and ordering; TRACE could assess whether the repair is adaptive or inefficient. Prefix metrics summarize the same example through failure at depth one, zero complete-trajectory executability, and terminal success despite that failure. Thus prefix metrics distinguish successful recovery after an invalid action from error-free execution, even when both trajectories receive the same terminal-success label.
\FloatBarrier
\end{revision}

\FloatBarrier
\section{Dual-Loop Validation Details}
\label{app:dual-loop}

\begin{figure}[!htbp]
\centering
\includegraphics[width=0.92\linewidth]{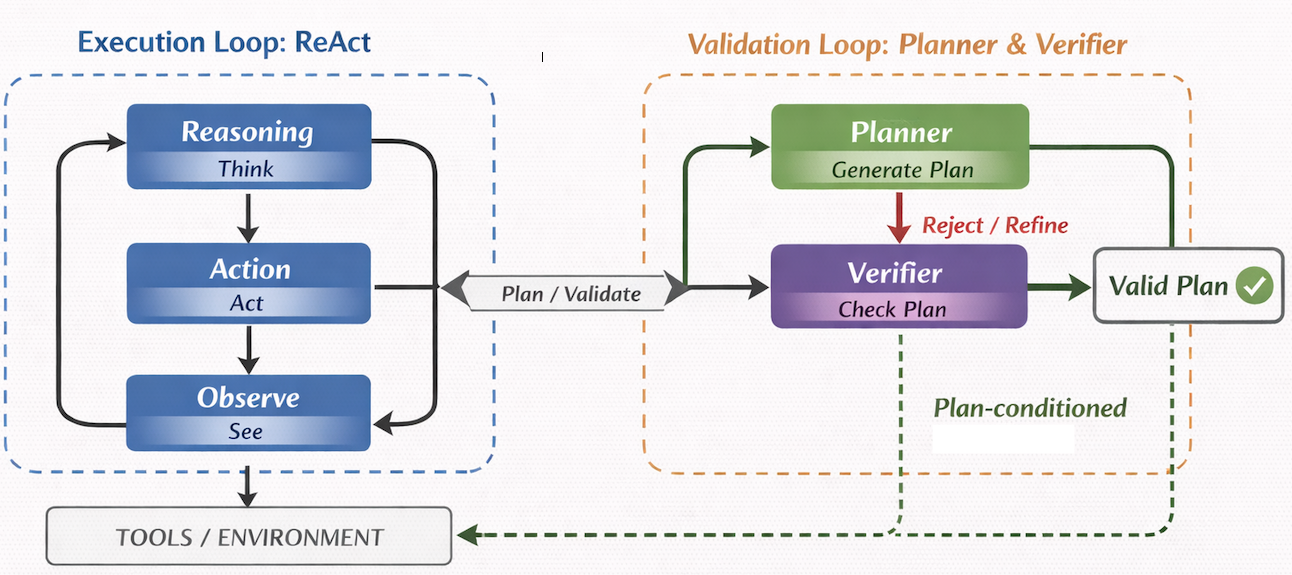}
\caption{Dual-Loop validation intervention. The planner proposes a structured candidate plan, the verifier either accepts it or returns targeted feedback, and the executor emits the only environment-facing actions scored by the prefix-level metrics.}
\label{fig:dual_loop}
\end{figure}

\rev{This appendix specifies the validation intervention in Section~\ref{sec:validation-intervention}. Only the evaluated executor trajectory, including the terminal no-tool action defined in Section~\ref{sec:prefix-metrics}, enters the prefix metrics. Planner and verifier messages influence subsequent behavior and cost but do not contribute to $T_i$. Figure~\ref{fig:dual_loop} summarizes the information flow.}

\subsection{Objects and Information Boundary}

Let $x$ denote the user goal, $\mathcal{U}$ the available tool specification, and $s_0$ a compact summary of benchmark-visible state before execution. Dual-Loop uses a planner $P$, verifier $V$, and executor $R$. At validation iteration $j$, the planner proposes a finite candidate plan
\[
    z^{(j)}=(z^{(j)}_1,\ldots,z^{(j)}_{M_j}),
\]
where each step describes an intended tool-level operation, argument values or constraints, and dependencies on prior planned steps, observations to be produced by those steps, or user-provided information. The plan length $M_j$ need not equal the eventual executed horizon $T_i$: the plan is conditioning context, not the object being scored.

The verifier maps $(x,\mathcal{U},s_0,z^{(j)})$ to either acceptance or a structured rejection record. The planner and verifier do not receive the benchmark reference trajectory, hidden final-success label, or future observations. After validation terminates, the executor interacts with the benchmark and emits
\[
    \hat{\tau}_i=(a_{i,1},\ldots,a_{i,T_i}).
\]
The evaluator scores this emitted trajectory directly. Thus, if the executor deviates from a verified plan, terminates early, or emits an invalid call, the metrics reflect the emitted actions exactly as they occurred.

\FloatBarrier

\subsection{Verifier Checklist}

The verifier is designed to reject candidate plans that contain predictable sources of non-executable prefixes before any environment action is taken. It applies the following local checklist, aligned with the first-failure taxonomy in \rev{Section~\ref{sec:first-failure-annotation}}:
\begin{enumerate}[leftmargin=*, nosep]
    \item \textbf{Tool and schema validity.} Each planned operation must refer to an available tool in $\mathcal{U}$, include required arguments, avoid unsupported arguments, and satisfy declared type or formatting constraints.
    \item \textbf{Dependency resolution.} Values referenced by a planned step must be available from the prompt, the initial state summary, or an earlier planned step. The verifier rejects forward references, stale bindings, and type-incompatible reuse.
    \item \textbf{Information availability.} Required user information or lookup results must either be present, requested before use, or represented as a dependency on an earlier planned step. This check catches plans that use information before collecting it.
    \item \textbf{Explicit preconditions.} If a tool requires prior authorization, a preceding state update, or a specific object to exist before use, the plan must satisfy that prerequisite before the tool is called.
    \item \textbf{State consistency.} Planned state mutations must be compatible with benchmark-visible state and with state changes induced by earlier planned steps. Examples include rejecting updates to absent objects or operations that consume unavailable resources.
\end{enumerate}
The checklist is intentionally local. It is not a proof of global optimality, and it does not replace benchmark replay after execution. The benchmark-side evaluator remains the authority for FED, TER, final success, and RR.

\subsection{Repair Loop and Iteration Budgets}

If the verifier rejects a candidate plan, its feedback identifies the earliest detected failing step, the failed check, and the information needed for repair when that information is available from $x$, $\mathcal{U}$, or $s_0$. The planner then proposes a revised candidate before any environment-facing action is emitted. \rev{In the BFCL-v3/ToolSandbox experiments, the validation loop} is capped at $K_{\mathrm{val}}=5$ candidate plans. If no candidate is accepted within this cap, the validation stage returns a failure flag and the executor receives no accepted plan as a strong constraint. The rollout is not discarded; any environment actions subsequently emitted by the executor are scored in the same way as all other rollouts. Cap exhaustion is therefore method-side behavior, not a prefix failure.

The validation cap is separate from the ReAct execution cap. \rev{For BFCL-v3 and ToolSandbox, we use $K_{\mathrm{exec}}=10$ ReAct action iterations} for the environment-facing executor. This separation is important because the two loops bound different resources: $K_{\mathrm{val}}$ limits unscored pre-execution search over candidate plans, whereas $K_{\mathrm{exec}}$ limits the scored action horizon available for multi-step tool use. A validation iteration does not increment $T_i$; only executor-emitted environment actions, or the terminal no-tool action described in Section~\ref{sec:prefix-metrics}, contribute to $T_i$ and to PEC, FED, TER, AUC-PEC, and RR.

\subsection{Metric Accounting and Controls}

The following conventions keep the intervention aligned with the prefix-level definitions:
\begin{enumerate}[leftmargin=*, nosep]
    \item Candidate plans, verifier decisions, and verifier feedback are not benchmark actions. They affect metrics only through the executor's subsequent environment-facing behavior.
    \item FED is computed after execution from $\hat{\tau}_i$. A verified plan that later leads the executor to an invalid call is counted as an ordinary prefix failure.
    \item RR is computed only for rollouts with $\mathrm{FED}_i\leq T_i$. Rejected candidate plans before execution are not RR-eligible because no benchmark prefix has failed.
    \item Validation does not modify benchmark state before execution. It can improve PEC, TER, or AUC-PEC only by changing the environment-facing actions emitted by the executor.
\end{enumerate}
\rev{Within each benchmark--model condition, all variants use the same task identifiers, tool specifications, executor horizon cap, and decoding rule. The Dual-Loop--Plan-Conditioned ReAct comparison measures the joint effect of validation, revision, and their additional inference cost under matched executor conditions. The scored trajectory semantics are the same across variants. Appendix~\ref{app:transfer} gives the verifier-sensitivity results and experimental controls.} Planner and verifier calls are outside the scored trajectory, but should be included when comparing token use, latency, or deployment cost.

\subsection{Limitations of the Intervention}

\rev{The verifier can miss hidden-state constraints, underspecified intent, or semantic requirements outside the tool interface, and can reject executable alternatives. Pre-execution validation also adds inference cost. The prefix metrics characterize how Dual-Loop's effects vary across environments and configurations (Appendix~\ref{app:transfer}).}

\FloatBarrier
\section{Additional Experimental Settings and Compute Resources}
\label{app:compute-resources}

\paragraph{Decoding and rollout settings.}
\rev{The following decoding settings and aggregate runtime apply to the 4,800-rollout BFCL-v3/ToolSandbox grid.} All LLM conditions used the least-stochastic provider-supported configuration.
For GPT-4.1 and Llama-4-Maverick-Instruct, we set temperature to $0$, used
top-$p=1$ when exposed, and disabled presence and frequency penalties. For the
GPT-5.4 reasoning condition, unsupported sampling controls were not set; instead
we fixed \texttt{reasoning.effort=high} and kept the remaining provider-exposed
generation controls at their defaults. To equalize generation budget, every
planner, verifier, and executor call used a $4{,}000$-token response cap; for
APIs that distinguish hidden reasoning tokens from visible text, this cap is
applied to the provider's total output-token budget. The environment-facing
ReAct execution loop was capped at $K_{\mathrm{exec}}=10$ tool-action iterations
for every ReAct-style agent variant, while Dual-Loop additionally used the
separate pre-execution planner--verifier budget $K_{\mathrm{val}}=5$ described
in \rev{Appendix~\ref{app:dual-loop}}; validation attempts are method-side computation
and are not counted as tool actions in $T_i$.

\paragraph{Compute resources.}
Our experiments evaluate agent rollouts rather than train or fine-tune models. All LLM inference was performed through hosted model APIs; the local machine only orchestrated benchmark execution, tool-state updates, trajectory logging, and metric computation. Model inference was served by provider-managed accelerator infrastructure behind API endpoints, so per-request GPU type, GPU count, and accelerator occupancy were not exposed to us. We therefore report local orchestration compute separately from hosted inference compute. \rev{The BFCL-v3/ToolSandbox grid contains} $200 \times 2 \times 3 \times 4 = 4{,}800$ rollouts across two datasets, three models, and four agent variants. A CPU-only local worker is sufficient for the evaluation harness; \rev{the BFCL-v3/ToolSandbox runs used a Linux worker} with 16 vCPUs and 64 GB RAM, with no local GPU allocated. Wall-clock time was dominated by remote model latency and API rate limits rather than local computation; using 16 asynchronous API-request workers, \rev{the 4,800-rollout BFCL-v3/ToolSandbox grid completed in approximately 72 hours}. \rev{Offline computation of PEC, FED, TER, AUC-PEC, RR, and bootstrap confidence intervals for this grid completed in under 1 hour on the same machine.}

\rev{The four-method, three-benchmark audit comprises 4,800 BFCL-v3/ToolSandbox rollouts and 1,968 Airline/Retail rollouts, totaling 6,768. The verifier-sensitivity study uses separate paired runs outside this total. The approximately 72-hour runtime applies only to BFCL-v3/ToolSandbox. Appendix~\ref{app:transfer} reports agent-side token and latency measurements for Airline/Retail and cost differences for the sensitivity study, excluding simulator and evaluator calls.}

\FloatBarrier
\begin{revision}
\section{Transfer Audit, Cost, and Verifier Sensitivity}
\label{app:transfer}

\subsection{Airline and Retail Audit Protocol}

We evaluate all $50$ Airline and $114$ Retail tasks in the base splits of
$\tau^3$-bench v1.0.1 in its text-based, turn-taking (half-duplex)
environment, using the benchmark's simulator protocol, task specifications,
and reward schema. This is complete coverage of these two base splits, not of every
$\tau^3$-bench domain. Each task is evaluated once under each of the four
methods and three model conditions. All methods are evaluated on the same tasks.
Table~\ref{tab:tau-results} pools the $164$ tasks, assigning equal weight
to each task rather than to each domain. RR is computed over the eligible
episodes in this pooled cohort.

For TER, mean FED, AUC-PEC, and final success, the intervals are $95\%$
task-level percentile bootstrap intervals from $10{,}000$ resamples.
RR uses a two-sided $95\%$ Wilson interval and reports the exact number of
recovery successes and eligible episodes. These intervals describe
task-level variation under the fixed rollout protocol; one rollout per
task does not estimate within-task variation across repeated runs.
Intervals for individual methods do not quantify uncertainty in
paired differences between methods. The following subsections report agent costs and a separate
sensitivity study with paired method differences.

\begin{table}[!htbp]
\centering
\footnotesize
\setlength{\tabcolsep}{3pt}
\renewcommand{\arraystretch}{1.15}
\newcommand{\transferci}[2]{\shortstack{#1\\{}[#2]}}
\begin{tabular*}{\linewidth}{@{\extracolsep{\fill}}lccccc@{}}
\toprule
Method & TER & Mean FED & AUC-PEC & RR (s/n) & Final success\\
\midrule
\multicolumn{6}{l}{GPT-5.4 (high)}\\
ReAct & \transferci{.95}{.91,.98} & \transferci{5.92}{5.53,6.31}
& \transferci{.95}{.91,.98} & \shortstack{.22 [.06,.55]\\(2/9)}
& \transferci{.43}{.35,.51}\\
Prompt & \transferci{.94}{.90,.98} & \transferci{5.73}{5.34,6.13}
& \transferci{.94}{.91,.98} & \shortstack{.10 [.02,.40]\\(1/10)}
& \transferci{.46}{.38,.54}\\
Plan-Cond. & \transferci{.93}{.88,.96} & \transferci{5.84}{5.45,6.24}
& \transferci{.93}{.89,.96} & \shortstack{.25 [.09,.53]\\(3/12)}
& \transferci{.51}{.43,.59}\\
Dual-Loop & \transferci{.93}{.89,.97} & \transferci{5.74}{5.36,6.14}
& \transferci{.93}{.89,.97} & \shortstack{.27 [.10,.57]\\(3/11)}
& \transferci{.46}{.38,.53}\\
\midrule
\multicolumn{6}{l}{GPT-4.1}\\
ReAct & \transferci{.84}{.79,.90} & \transferci{6.47}{5.98,6.99}
& \transferci{.91}{.87,.94} & \shortstack{.54 [.35,.71]\\(14/26)}
& \transferci{.46}{.39,.54}\\
Prompt & \transferci{.82}{.76,.88} & \transferci{6.23}{5.79,6.68}
& \transferci{.89}{.84,.93} & \shortstack{.45 [.28,.62]\\(13/29)}
& \transferci{.51}{.43,.59}\\
Plan-Cond. & \transferci{.79}{.73,.85} & \transferci{6.20}{5.72,6.70}
& \transferci{.87}{.82,.91} & \shortstack{.41 [.26,.58]\\(14/34)}
& \transferci{.49}{.41,.57}\\
Dual-Loop & \transferci{.81}{.75,.87} & \transferci{6.29}{5.54,7.33}
& \transferci{.86}{.81,.91} & \shortstack{.48 [.32,.65]\\(15/31)}
& \transferci{.45}{.37,.52}\\
\midrule
\multicolumn{6}{l}{Llama-4 Maverick}\\
ReAct & \transferci{.52}{.44,.59} & \transferci{2.66}{2.37,2.98}
& \transferci{.56}{.49,.63} & \shortstack{.11 [.06,.20]\\(9/79)}
& \transferci{.18}{.12,.24}\\
Prompt & \transferci{.55}{.48,.63} & \transferci{2.95}{2.63,3.29}
& \transferci{.60}{.52,.67} & \shortstack{.12 [.07,.22]\\(9/73)}
& \transferci{.18}{.12,.24}\\
Plan-Cond. & \transferci{.60}{.53,.68} & \transferci{2.71}{2.46,2.98}
& \transferci{.65}{.58,.72} & \shortstack{.06 [.02,.15]\\(4/65)}
& \transferci{.18}{.12,.24}\\
Dual-Loop & \transferci{.51}{.43,.59} & \transferci{2.56}{2.24,2.93}
& \transferci{.54}{.47,.61} & \shortstack{.06 [.03,.14]\\(5/80)}
& \transferci{.14}{.09,.20}\\
\bottomrule
\end{tabular*}
\caption{Complete Airline/Retail base-split audit. Brackets are
$95\%$ intervals. For RR, s/n is the exact recovery-success/eligible count.
Prompt denotes Prompt Planning; Plan-Cond.\ denotes Plan-Conditioned
ReAct. All values are empirical estimates.}
\label{tab:tau-results}
\end{table}
\FloatBarrier
\subsection{Agent-Side Token Use and Latency}
\label{app:transfer-cost}

Table~\ref{tab:tau-cost} reports mean agent token use and cumulative
request latency per episode for the Airline/Retail audit. Agent tokens sum provider-reported input and output
tokens across planner, verifier, planner-revision, and executor requests.
Recorded reasoning tokens are included in output tokens and counted
once. Latency is the client-observed cumulative latency of those same
requests, averaged per episode. Simulator and evaluator requests are
excluded from both quantities. These latency values measure cumulative
agent-call time rather than complete benchmark wall-clock duration.

\begin{table}[!htbp]
\centering
\small
\setlength{\tabcolsep}{4pt}
\renewcommand{\arraystretch}{1.3}
\begin{tabular*}{\linewidth}{@{\extracolsep{\fill}}lcccc@{}}
\toprule
Model & ReAct & Prompt & Plan-Cond. & Dual-Loop\\
\midrule
GPT-5.4 (high) & 86k / 57s & 94k / 68s & 95k / 68s & 98k / 83s\\
GPT-4.1 & 142k / 65s & 156k / 70s & 161k / 73s & 180k / 78s\\
Llama-4 Maverick & 374k / 61s & 409k / 66s & 412k / 67s & 399k / 66s\\
\bottomrule
\end{tabular*}
\caption{Mean agent tokens / mean cumulative agent-call latency
on the Airline/Retail audit; k denotes thousands of tokens and s denotes
seconds. Prompt and Plan-Cond.\ abbreviate Prompt Planning and
Plan-Conditioned ReAct.}
\label{tab:tau-cost}
\end{table}

Relative to ReAct, Dual-Loop changes mean tokens/latency by approximately
$+14\%/+46\%$ for GPT-5.4, $+27\%/+20\%$ for GPT-4.1, and
$+7\%/+8\%$ for Llama-4 Maverick. Relative to Plan-Conditioned ReAct,
the corresponding approximate differences are
$+3\%/+22\%$, $+12\%/+7\%$, and $-3\%/-1\%$.

The Dual-Loop--Plan-Conditioned contrast holds the model condition,
executor, task identifiers, tool specification, decoding rule, and
execution horizon fixed while adding validation and revision. These
are net episode-level cost differences: verifier and revision costs
are combined with changes in downstream execution cost. The comparison
varies validation behavior and inference budget together. Token totals
should be compared within a model condition because tokenization and
reasoning-token accounting differ across providers and models.

Per-method token-use and latency measurements are available for the
Airline/Retail audit. For BFCL-v3 and ToolSandbox, compute reporting is
limited to aggregate runtime; the Airline/Retail measurements do not
establish their per-method costs.

\FloatBarrier
\subsection{Matched Verifier Sensitivity}
\label{app:verifier-sensitivity}

A separate sensitivity study evaluates the same $164$ Airline/Retail
tasks with a GPT-5.4-high planner and executor. For each task, all
experimental conditions use the same initial plan, tool specifications,
and execution horizon. Each condition's cost includes the generation of
the shared initial plan once, together with its verification, plan-revision,
and execution requests; simulator and evaluator requests are excluded.
Differences in Table~\ref{tab:verifier-sensitivity} are measured against
a separately evaluated, task-paired Plan-Conditioned ReAct baseline.
This baseline belongs to the sensitivity experiment; it is distinct from
the Plan-Conditioned ReAct estimates in Tables~\ref{tab:tau-results}
and~\ref{tab:tau-cost}.

The $95\%$ intervals for metric differences use $100{,}000$ paired
task-bootstrap resamples, preserving the task pairing between each
condition and its baseline. These characterize variation across tasks
under paired resampling and do not measure within-task variability over
repeated runs. Cost
differences are reported as point estimates without intervals.

\begin{table}[!htbp]
\centering
\footnotesize
\setlength{\tabcolsep}{2pt}
\renewcommand{\arraystretch}{1.3}
\newcommand{\sensitivityci}[2]{\shortstack{#1\\{}[#2]}}
\begin{tabular*}{\linewidth}{@{\extracolsep{\fill}}lcccc@{}}
\toprule
Configuration & $\Delta$TER & $\Delta$AUC-PEC & $\Delta$Final &
\shortstack{$\Delta$Tokens /\\$\Delta$Latency}\\
\midrule
\shortstack[l]{Deterministic\\acceptance}
& \sensitivityci{$+.000$}{$-.018,+.018$}
& \sensitivityci{$-.003$}{$-.018,+.008$}
& \sensitivityci{$-.073$}{$-.140,-.012$}
& $-8.4$k / $-3.2$s\\
\shortstack[l]{GPT-5.4\\$K_{\mathrm{val}}=1$}
& \sensitivityci{$-.006$}{$-.030,+.012$}
& \sensitivityci{$-.010$}{$-.030,+.005$}
& \sensitivityci{$-.012$}{$-.079,+.055$}
& $+0.7$k / $+43.1$s\\
\shortstack[l]{GPT-4.1\\$K_{\mathrm{val}}=5$}
& \sensitivityci{$+.000$}{$-.018,+.018$}
& \sensitivityci{$+.000$}{$-.007,+.008$}
& \sensitivityci{$-.085$}{$-.165,-.006$}
& $+7.1$k / $+30.9$s\\
\shortstack[l]{GPT-5.4\\$K_{\mathrm{val}}=5$}
& \sensitivityci{$+.006$}{$-.018,+.030$}
& \sensitivityci{$+.006$}{$-.015,+.030$}
& \sensitivityci{$+.000$}{$-.073,+.073$}
& $+40.7$k / $+176.9$s\\
\bottomrule
\end{tabular*}
\caption{Matched verifier-sensitivity differences against the
paired Plan-Conditioned ReAct rerun, with $95\%$ paired task-bootstrap
intervals. Model names identify the verifier; the planner and executor
remain GPT-5.4-high. The deterministic-acceptance condition makes no verifier
call and changes the executor's instructions.}
\label{tab:verifier-sensitivity}
\end{table}

All TER and AUC-PEC intervals include zero, with small point-estimate
changes in this environment and configuration. Equivalence was not
tested. The final-success intervals for deterministic acceptance and
the GPT-4.1 verifier are negative relative to the paired baseline;
the other two final-success intervals include zero. These paired
measurements quantify prefix reliability, terminal outcomes, and
inference cost separately for each verifier configuration.

\paragraph{Interpretation of verifier contrasts.}
Both deterministic acceptance and the Plan-Conditioned ReAct baseline
omit verifier calls, but deterministic acceptance tells the executor
that the plan was accepted without evaluation. Its contrast with the
baseline therefore changes executor instructions and does not measure
the effect of removing verification. Comparisons involving LLM verifiers
also change inference cost. The reported differences are specific to
the evaluated environment and configurations and do not isolate verification
logic from executor instructions and computational budget.
\FloatBarrier
\end{revision}

\FloatBarrier
\section{Depth-Wise PEC Checkpoints}
\label{app:pec-depths}
\rev{Table~\ref{tab:pec_depths}} reports checkpoint values for the fixed-cohort PEC curves used in the main analysis. These checkpoints are included for auditability and to show where along the tool-call trajectory the methods begin to separate. Because $\widehat{\mathrm{PEC}}(k)$ is a survival estimate over the same rollout cohort, all $N=200$ rollouts in a dataset--model--method block remain in the denominator; the values should be compared within blocks and interpreted as prefix-reliability diagnostics, not as eligible-depth conditional accuracy or path-efficiency scores. Early depths are most informative for method comparison, while later depths document tail behavior where decreases can also reflect trajectories that terminate before depth $k$.

\rev{Equation~\eqref{eq:pec-support} expresses fixed-cohort PEC as the fraction of rollouts reaching depth $k$, multiplied by the fraction of those rollouts with executable prefixes. Table~\ref{tab:pec_depths} reports their product; it does not separately report depth support and eligible-depth PEC.}

\begin{table}[!htbp]
\centering
\scriptsize
\setlength{\tabcolsep}{3.2pt}
\begin{tabular}{@{}lllcccc@{}}
\toprule
Dataset & LLM & Method & PEC@1 & PEC@2 & PEC@3 & PEC@4 \\
\midrule
BFCL & GPT-5.4 (high) & ReAct & \shortstack[c]{0.95\\{\scriptsize[0.93,0.98]}} & \shortstack[c]{0.90\\{\scriptsize[0.85,0.94]}} & \shortstack[c]{0.81\\{\scriptsize[0.75,0.86]}} & \shortstack[c]{0.68\\{\scriptsize[0.61,0.74]}} \\
 &  & Prompt Planning & \shortstack[c]{0.96\\{\scriptsize[0.93,0.98]}} & \shortstack[c]{0.88\\{\scriptsize[0.83,0.93]}} & \shortstack[c]{0.82\\{\scriptsize[0.77,0.88]}} & \shortstack[c]{0.70\\{\scriptsize[0.64,0.77]}} \\
 &  & Plan-Cond. ReAct & \shortstack[c]{0.77\\{\scriptsize[0.70,0.82]}} & \shortstack[c]{0.61\\{\scriptsize[0.55,0.69]}} & \shortstack[c]{0.49\\{\scriptsize[0.42,0.56]}} & \shortstack[c]{0.42\\{\scriptsize[0.36,0.49]}} \\
 &  & Dual-Loop & \shortstack[c]{0.98\\{\scriptsize[0.96,0.99]}} & \shortstack[c]{0.97\\{\scriptsize[0.94,0.99]}} & \shortstack[c]{0.94\\{\scriptsize[0.91,0.97]}} & \shortstack[c]{0.82\\{\scriptsize[0.77,0.88]}} \\
 & GPT-4.1 & ReAct & \shortstack[c]{0.93\\{\scriptsize[0.89,0.96]}} & \shortstack[c]{0.84\\{\scriptsize[0.79,0.89]}} & \shortstack[c]{0.74\\{\scriptsize[0.68,0.80]}} & \shortstack[c]{0.61\\{\scriptsize[0.55,0.68]}} \\
 &  & Prompt Planning & \shortstack[c]{0.97\\{\scriptsize[0.94,0.99]}} & \shortstack[c]{0.93\\{\scriptsize[0.89,0.96]}} & \shortstack[c]{0.86\\{\scriptsize[0.81,0.91]}} & \shortstack[c]{0.71\\{\scriptsize[0.65,0.77]}} \\
 &  & Plan-Cond. ReAct & \shortstack[c]{0.65\\{\scriptsize[0.57,0.71]}} & \shortstack[c]{0.48\\{\scriptsize[0.41,0.55]}} & \shortstack[c]{0.39\\{\scriptsize[0.32,0.46]}} & \shortstack[c]{0.33\\{\scriptsize[0.26,0.39]}} \\
 &  & Dual-Loop & \shortstack[c]{0.99\\{\scriptsize[0.98,1.00]}} & \shortstack[c]{0.97\\{\scriptsize[0.95,0.99]}} & \shortstack[c]{0.95\\{\scriptsize[0.93,0.98]}} & \shortstack[c]{0.82\\{\scriptsize[0.77,0.88]}} \\
 & Llama-4 Maverick & ReAct & \shortstack[c]{0.89\\{\scriptsize[0.84,0.93]}} & \shortstack[c]{0.71\\{\scriptsize[0.66,0.78]}} & \shortstack[c]{0.59\\{\scriptsize[0.53,0.67]}} & \shortstack[c]{0.46\\{\scriptsize[0.39,0.53]}} \\
 &  & Prompt Planning & \shortstack[c]{0.94\\{\scriptsize[0.91,0.97]}} & \shortstack[c]{0.88\\{\scriptsize[0.82,0.92]}} & \shortstack[c]{0.80\\{\scriptsize[0.74,0.85]}} & \shortstack[c]{0.65\\{\scriptsize[0.58,0.71]}} \\
 &  & Plan-Cond. ReAct & \shortstack[c]{0.53\\{\scriptsize[0.46,0.59]}} & \shortstack[c]{0.23\\{\scriptsize[0.18,0.29]}} & \shortstack[c]{0.10\\{\scriptsize[0.06,0.14]}} & \shortstack[c]{0.05\\{\scriptsize[0.02,0.08]}} \\
 &  & Dual-Loop & \shortstack[c]{0.98\\{\scriptsize[0.96,1.00]}} & \shortstack[c]{0.94\\{\scriptsize[0.91,0.97]}} & \shortstack[c]{0.90\\{\scriptsize[0.85,0.94]}} & \shortstack[c]{0.78\\{\scriptsize[0.72,0.83]}} \\
\midrule
ToolSandbox & GPT-5.4 (high) & ReAct & \shortstack[c]{0.92\\{\scriptsize[0.88,0.95]}} & \shortstack[c]{0.81\\{\scriptsize[0.75,0.86]}} & \shortstack[c]{0.59\\{\scriptsize[0.52,0.66]}} & \shortstack[c]{0.47\\{\scriptsize[0.40,0.54]}} \\
 &  & Prompt Planning & \shortstack[c]{0.94\\{\scriptsize[0.91,0.97]}} & \shortstack[c]{0.88\\{\scriptsize[0.83,0.93]}} & \shortstack[c]{0.63\\{\scriptsize[0.56,0.69]}} & \shortstack[c]{0.56\\{\scriptsize[0.48,0.62]}} \\
 &  & Plan-Cond. ReAct & \shortstack[c]{0.67\\{\scriptsize[0.60,0.73]}} & \shortstack[c]{0.49\\{\scriptsize[0.42,0.56]}} & \shortstack[c]{0.35\\{\scriptsize[0.29,0.42]}} & \shortstack[c]{0.32\\{\scriptsize[0.26,0.39]}} \\
 &  & Dual-Loop & \shortstack[c]{0.97\\{\scriptsize[0.95,0.99]}} & \shortstack[c]{0.95\\{\scriptsize[0.92,0.98]}} & \shortstack[c]{0.69\\{\scriptsize[0.62,0.75]}} & \shortstack[c]{0.62\\{\scriptsize[0.56,0.69]}} \\
 & GPT-4.1 & ReAct & \shortstack[c]{0.94\\{\scriptsize[0.91,0.97]}} & \shortstack[c]{0.81\\{\scriptsize[0.75,0.86]}} & \shortstack[c]{0.58\\{\scriptsize[0.52,0.65]}} & \shortstack[c]{0.45\\{\scriptsize[0.38,0.52]}} \\
 &  & Prompt Planning & \shortstack[c]{0.96\\{\scriptsize[0.93,0.98]}} & \shortstack[c]{0.90\\{\scriptsize[0.85,0.94]}} & \shortstack[c]{0.64\\{\scriptsize[0.56,0.70]}} & \shortstack[c]{0.56\\{\scriptsize[0.48,0.62]}} \\
 &  & Plan-Cond. ReAct & \shortstack[c]{0.55\\{\scriptsize[0.47,0.61]}} & \shortstack[c]{0.37\\{\scriptsize[0.30,0.44]}} & \shortstack[c]{0.27\\{\scriptsize[0.21,0.34]}} & \shortstack[c]{0.24\\{\scriptsize[0.18,0.30]}} \\
 &  & Dual-Loop & \shortstack[c]{0.99\\{\scriptsize[0.97,1.00]}} & \shortstack[c]{0.96\\{\scriptsize[0.94,0.99]}} & \shortstack[c]{0.69\\{\scriptsize[0.63,0.76]}} & \shortstack[c]{0.64\\{\scriptsize[0.57,0.70]}} \\
 & Llama-4 Maverick & ReAct & \shortstack[c]{0.87\\{\scriptsize[0.82,0.92]}} & \shortstack[c]{0.68\\{\scriptsize[0.60,0.74]}} & \shortstack[c]{0.45\\{\scriptsize[0.38,0.52]}} & \shortstack[c]{0.29\\{\scriptsize[0.23,0.36]}} \\
 &  & Prompt Planning & \shortstack[c]{0.92\\{\scriptsize[0.88,0.95]}} & \shortstack[c]{0.80\\{\scriptsize[0.74,0.85]}} & \shortstack[c]{0.53\\{\scriptsize[0.46,0.60]}} & \shortstack[c]{0.41\\{\scriptsize[0.34,0.47]}} \\
 &  & Plan-Cond. ReAct & \shortstack[c]{0.36\\{\scriptsize[0.30,0.43]}} & \shortstack[c]{0.10\\{\scriptsize[0.06,0.14]}} & \shortstack[c]{0.09\\{\scriptsize[0.06,0.13]}} & \shortstack[c]{0.07\\{\scriptsize[0.04,0.12]}} \\
 &  & Dual-Loop & \shortstack[c]{0.99\\{\scriptsize[0.97,1.00]}} & \shortstack[c]{0.88\\{\scriptsize[0.83,0.93]}} & \shortstack[c]{0.65\\{\scriptsize[0.58,0.71]}} & \shortstack[c]{0.55\\{\scriptsize[0.48,0.62]}} \\
\bottomrule
\end{tabular}

\caption{Depth-wise prefix executability for the $200$-instance BFCL multi-turn sample and the $200$-instance ToolSandbox sample under three LLM conditions. All entries are empirical estimates. Bracketed values are $95\%$ percentile confidence intervals from $10{,}000$ task-level bootstrap resamples. PEC@$k$ uses the fixed-cohort survival estimate in Eq.~\eqref{eq:pec-estimator}: the numerator counts rollouts with $\mathrm{FED}>k$, and the denominator is all evaluated rollouts in the dataset-LLM-method block.}
\label{tab:pec_depths}
\end{table}

\FloatBarrier

\end{document}